\documentclass[conference]{IEEEtran}
\usepackage{cite}
\usepackage{amsmath,amssymb,amsfonts}
\usepackage{algorithmic}
\usepackage{graphicx}
\usepackage{textcomp}
\usepackage{xcolor}
\usepackage{float}
\usepackage{hyperref}
\usepackage[capitalize]{cleveref}
\usepackage{comment}
\usepackage{array}

\usepackage{flushend}

\newcommand{\citet}{\cite}

\DeclareMathOperator*{\argmin}{arg\,min}

\usepackage{xcolor}
\usepackage{soul}

\usepackage{caption}
\usepackage{subcaption}

\usepackage{booktabs}
\usepackage{multirow}

\def\BibTeX{{\rm B\kern-.05em{\sc i\kern-.025em b}\kern-.08em
    T\kern-.1667em\lower.7ex\hbox{E}\kern-.125emX}}

\begin{document}

\title{Bayesian Optimization for auto-tuning GPU kernels}


\author{\IEEEauthorblockN{Floris-Jan Willemsen}
\IEEEauthorblockA{\textit{Netherlands eScience Center}\\ \textit{University of Amsterdam}\\
Amsterdam, the Netherlands \\
ORCID: 0000-0003-2295-8263}
\and
\IEEEauthorblockN{Rob van Nieuwpoort}
\IEEEauthorblockA{\textit{Netherlands eScience Center}\\ \textit{University of Amsterdam}\\
Amsterdam, the Netherlands \\
ORCID: 0000-0002-2947-9444}
\and
\IEEEauthorblockN{Ben van Werkhoven}
\IEEEauthorblockA{\textit{Netherlands eScience Center}\\
Amsterdam, the Netherlands \\
ORCID: 0000-0002-7508-3272}
}

\maketitle
\thispagestyle{plain}
\pagestyle{plain}

\begin{abstract}
Finding optimal parameter configurations for tunable GPU kernels is a non-trivial exercise for large search spaces, even when automated. 
This poses an optimization task on a non-convex search space, using an expensive to evaluate function with unknown derivative. 
These characteristics make a good candidate for Bayesian Optimization, which has not been applied to this problem before. 

However, the application of Bayesian Optimization to this problem is challenging. We demonstrate how to deal with the rough, discrete, constrained search spaces, containing invalid configurations. 
We introduce a novel contextual variance exploration factor, as well as new  acquisition functions with improved scalability, combined with an informed acquisition function selection mechanism. 

By comparing the performance of our Bayesian Optimization implementation on various test cases to the existing search strategies in Kernel Tuner, as well as other Bayesian Optimization implementations, we demonstrate that our search strategies generalize well and consistently outperform other search strategies by a wide margin. 
\end{abstract}


\section{Introduction}

Graphics Processing Units (GPUs) have revolutionized the computing landscape in the past decade~\cite{heldens2020landscape}, and are seen as one of the enabling factors in recent breakthroughs in Artificial Intelligence~\cite{lecun2015deep}. While GPUs are used to enable scientific computing workloads in many fields, including climate modeling, artificial intelligence, and quantum physics, it is actually very hard to unlock the full computational power of the GPU~\cite{werkhoven2020lessons}.

There are many different design choices that have a strong and hard to predict impact on the performance of GPU kernels, including the way that a computation is parallelized and mapped on the thread blocks and individual threads. Other design choices include different algorithms to implement certain parts of the computation, or what data types and data layouts to use in the various memory spaces available to GPU applications.
This is also true for many code optimizations that may or may not be applied, and many of which also introduce parameters with different sets of possible values, such as tiling factors, vector data types, or partial loop unrolling factors. GPU kernels also have a number inherent parameters in terms of the number of thread blocks and the number of threads per block that are used to execute the kernel.

Specific to optimizing GPU kernels is the fact that a great number of combinations of parameter values in the search space result in kernel configurations that cannot be compiled or executed on the GPU. For example, because the configuration requires more shared memory than available to a single thread block, or because the combination of thread block dimensions and register usage per thread exceeds the size of the register file on the GPU, or because of implementation-specific constraints in the kernel. 
The multitude of implementation choices and invalid configurations for GPU kernels result in very large, non-convex, and discontinuous kernel design spaces~\cite{ryoo_program_2008, nugteren_cltune_2015, spafford_maestro_2010, lim_autotuning_2017}.

Several researchers have suggested using machine learning or metaheuristics to speedup the search process~\cite{kisuki2003combined, seymour2008comparison, balaprakash2011can, ansel_opentuner_2014, nugteren2015cltune, filipovivc2017autotuning, KernelTuner}. However, several of these approaches have been shown to not outperform random search~\cite{nugteren2015cltune, balaprakash2011can}, or do not have a clear winner that clearly outperforms other methods and reliably returns highly-efficient configurations for a wide range of different kernels and devices~\cite{filipovivc2017autotuning, KernelTuner}.
%
Therefore, in this paper, we explore the application of Bayesian Optimization (abbreviated to \textit{BO}) within the problem domain of auto-tuning a GPU kernel to a particular GPU. BO is known as an optimization method capable of quickly and reliably finding the global optimum of functions with unknown derivatives.

In this paper, we present the following contributions:
\begin{itemize}
    \item We present an implementation of a Bayesian Optimization search strategy, with a customized search space representation, dynamic exploration factor, and novel acquisition functions, specifically designed for auto-tuning GPU kernels.
    \item We demonstrate that our implementation of BO outperforms other existing search strategies in Kernel Tuner~\cite{KernelTuner} and existing widely-used BO frameworks by a wide margin.
    On average, our best implementation performs 49.7\% better at finding parameter configurations than the best other search strategy in Kernel Tuner, and 75\% better than the second-best. 
    \item We integrate our BO implementation in the open source Kernel Tuner, making it directly available for use. Additionally, we extend Kernel Tuner with a simulation mode, to enable benchmarking of search strategies without the need for a GPU. 
\end{itemize}

\cref{sec:related-work} discusses the related works. 
\cref{sec:application-implementation} describes the application and optimization of Bayesian Optimization to the problem domain.
Section~\ref{sec:results} presents our experimental results, and Section~\ref{sec:conclusions} concludes.

\section{Related Work}\label{sec:related-work}
From chocolate chip cookies~\cite{BO_cookies} to COVID~\cite{BO_COVID}: Bayesian Optimization has been applied in a wide range of subjects, including auto-tuning. To grasp the current state of BO in GPU auto-tuning, this section briefly explores related applications of BO in auto-tuning. For further reference,~\citet{BO_review} includes a review of BO and possible applications, listing a wide range of areas where BO has been applied. 

The parameters of high-performance libraries like Basic Linear Algebra Subroutines (\textit{BLAS}) used to be manually tuned and hard-coded, and at runtime selected based on architecture and some input characteristics. Recent developments require a more dynamic approach.~\citet{decision_trees_GPU_BLAS_library} applies decision trees to tackle this problem, showing significant performance gains on NVIDIA and ARM SoC GPUs. 
BLAS-libraries offer a clearly defined problem, as opposed to the unknown kernels inserted by the user in our case. Nevertheless, this shows that predictive modelling methods can be useful for GPU auto-tuning.  

Another example of application of a machine learning algorithm to auto-tuning is Adaptive Sampling, a form of Active Learning, in Adaptive Sampling Kit (\textit{ASK})~\cite{ASK}. 
Like BO, Adaptive Sampling revolves around optimizing the sampling process by focusing on areas of uncertainty. However, Active Learning in general differs from BO as it has different selection mechanisms and is most often used for classification~\cite{Active_Learning_vs_BO}. 


In a recent study, Wu et al.~\citet{CPU_PolyBench_BO} apply BO to a search space of LLVM Clang / Polly pragma configurations on the PolyBench Benchmark suite. Their method appears to be most successful with random forests. 
Although specifically targeting the Polybench benchmarks and CPU auto-tuning, this shows that optimization of discrete performance search spaces with BO can be successful. 


An interesting approach is the notion of Structured Bayesian Optimization (\textit{SBO}), in which expert knowledge (such as performance models) are combined with BO to give the algorithm an edge over generic BO. This concept of SBO made its debut in auto-tuning with a publication by~\citet{BOAT} concerning the BespOke Auto-Tuner (\textit{BOAT}), which provides custom-tailored auto-tuners based on expert knowledge. 
Although this structured BO is shown to be effective and mitigates the reportedly poor performance of BO on a high number of dimensions, it is intended for clearly specified problems where expert knowledge can be obtained and applied (e.g. BLAS), and as such does not generalize to our problem domain. 

Miyazaki et al.~\citet{Miyazaki_HPC_BO} apply BO to increase the energy efficiency of HPC systems. 
Their case concerns the tuning of a HPC system to improve its ranking on the GREEN 500 benchmark list. As such, the right balance between performance and energy consumption must be found.
While their approach is geared specifically towards the GREEN 500 benchmark, their positive results show potential for the application of BO to a broader problem domain. 


Finally, Liu et al.~\citet{GPTune} propose GPTune, a framework specifically targeting exascale systems using modified Gaussian Processes for multi-task auto-tuning of applications, allowing for function evaluation in parallel. 
Although GPTune focuses on exascale applications instead of GPU auto-tuning, it demonstrates the potential of parallel multi-objective BO on discrete search spaces. 

While the Bayesian Optimization applications seen in this related works section relate to auto-tuning and approaches contain elements similar to our problem domain, to the best of our knowledge, BO has not yet been applied to GPU auto-tuning. 
The GPU auto-tuning problem domain is different from CPU auto-tuning. CPU auto-tuning focuses mostly on compiler flags, while GPU auto-tuning also deals with resource partitioning, strictly invalid configurations (configurations resulting in errors instead of a performance impact), and specific constraints (e.g., a fitting block size for matrix multiplication).

\section{Design and Implementation}\label{sec:application-implementation}

In this section, we describe the application of Bayesian Optimization to the problem domain of auto-tuning GPU kernels. We start with a brief description of BO in \cref{sec:bayesian-optimization}, and \cref{sec:design-implementation-implementation} describes our implementation. 

\subsection{Bayesian Optimization}\label{sec:bayesian-optimization}

Essentially Bayesian Optimization comprises an objective function, a surrogate model and an acquisition function. 
The objective function $f: \mathcal{X} \rightarrow \mathbb{R}$ is considered as an otherwise unknown (black box) function for which the global optimum is to be found. As the goal of auto-tuning is usually to minimize the usage of resources (such as energy consumption or time), the optimum will from here on be assumed to be a minimum. This global optimum is then expressed by $x' = \argmin\limits_{x \epsilon \mathcal{X}} f(x)$, where $\mathcal{X}$ is the search space, consisting of all possible combinations of all parameter values. 
The goal of the \emph{surrogate model} is to emulate the objective function, especially in the optimization direction. 
The \emph{acquisition function} then optimizes the surrogate model to suggest the next parameter configuration for evaluation with the objective function. 
The resulting observation is fitted to the surrogate model, improving the likeness of the surrogate model to the objective function, making the next suggestion of the acquisition function more precise.

\subsection{Gaussian processes \& covariance functions}\label{sec:gaussian-processes}

The surrogate model can be formed with various techniques, such as Gaussian processes, Tree Parzen Estimators~\cite{hyperopt_tpe}, Support Vector Machines~\cite{SVM_regression_surrogate_model}, or Random Forests~\cite{gaussian_processes_vs_random_forest}. 
Each of these has their advantages and disadvantages. SVMs do not offer a probabilistic interpretation, Gaussian processes are computationally intensive, Random Forests tend to require more observations than Gaussian processes~\cite{gaussian_processes_vs_random_forest}, and TPEs do not model the interaction between hyperparameters. 

To form the surrogate model, we employ the commonly used Gaussian processes (\textit{GP}), which consist of a mean function and a covariance function. This covariance (or kernel function) describes the covariance of the random variables in the Gaussian process. While the mean function is usually a zero function, the choice of a kernel for the covariance function can have a significant impact on the surrogate model. 

Kernels come in several variations, such as linear and periodic kernels, which can be combined to form a custom kernel befitting a problem domain where there can be strong assumptions on the shape of the objective function given expert knowledge. 
A commonly used default kernel for BO is the Radial Basis Function (\textit{RBF}, also known as the Squared Exponential kernel), given several properties that make it an adequate default: it is universal, every function in the prior has infinitely many derivatives and it has two parameters, the lengthscale $l$ and signal or output variance $\sigma^2$~\cite{cov_functions_gp_kernels}. The lengthscale determines the smoothness of the function, whereas the output variance determines the scaling. 
A popular variant on the RBF kernel is the Radial Quadratic kernel, which is in essence a combination of RBFs with varying lengthscales, generally known as a \textit{scale mixture}. An additional parameter $\alpha$ determines the weighting of variation in scales. 

The Matérn kernel~\citet{matern_class_stein} 
can be seen as a generalization of RBF, which is more suitable for physical processes, as the Matérn kernel is finitely differentiable. 
In addition, Le et al.~\citet{fast_kernel_expansions_rbf_matern} have shown the Matérn kernel to be more suitable for high-dimensional inputs compared to the RBF kernel, as it suffers less from \textit{concentration of measurement} problems.
The Matérn kernel is defined by~\citet{cov_functions_gp_kernels} as $k(r) = \frac{2^{1-\nu}}{\Gamma(\nu)} (\frac{\sqrt{2\nu r}}{l})^\nu K_{\nu} (\frac{\sqrt{2\nu r}}{l})$, where $K_{\nu}$ is a modified Bessel function~\cite{bessel_function_abramowitz}, and $\nu$ and $l$ are positive parameters. 

The Matérn kernel contains a specific subclass of kernels: when the parameter $\nu$ is a positive half-integer, the Matérn kernel is simplified to the product of an exponential and a polynomial. According to~\citet{cov_functions_gp_kernels}, for $\nu \leq \frac{1}{2}$, the resulting process is very rough, whereas for $\nu \geq \frac{7}{2}$, the process becomes too smooth to model physical processes, making it practically indistinguishable from RBF. 
As such, $\nu = \frac{3}{2}$ and $\nu = \frac{5}{2}$ are the most interesting instances, which can be simplified to the following kernel functions respectively: \\ $k_{\nu=\frac{3}{2}}(r) = \Big(1 + \frac{\sqrt{3r}}{l}\Big) \exp\Big(-\frac{\sqrt{3r}}{l}\Big)$, and \\ $k_{\nu=\frac{5}{2}}(r) = \Big(1 + \frac{\sqrt{5r}}{l} +\frac{5r^2}{3l^2}\Big) \exp\Big(-\frac{\sqrt{5r}}{l}\Big)$. 

Choosing an appropriate covariance function is important to achieve good predictive performance within few evaluations~\citet{practical_BO}. 
For auto-tuning GPU kernels, we require a generally applicable covariance function as the objective function depends on the user input, simple covariance functions, such as linear and periodic, or combinations of these are not suited. 
While both RBF and Radial Quadratic are often used as defaults, they may not be particularly suited to our problem. These kernels tend to work best on smooth functions. As the lengthscale is usually adjusted based on the least smooth part of the surrogate model, discontinuities in the search space can severely disrupt the model. We thus use a fixed lengthscale. 

As such, the Matérn kernel, given the discrete method proposed, the rougher Matérn $\nu=\frac{3}{2}$ combined with a larger ($>1$) lengthscale appears to be a good choice to generalize well, making for a rougher function with a smoothing lengthscale. Alternatively $\nu=\frac{5}{2}$ with a shorter lengthscale ($<1$) may be a suitable choice as well, as a smoother function with a rougher lengthscale.

\subsection{Acquisition functions}\label{sec:acquisition-functions}

Having established a GP-surrogate model $g$, this must now be sequentially refined to approach the objective function $f$. As evaluation of $f$ is expensive, there must be a clever approach to which points to refine. To obtain the most useful information from a single evaluation, either the most uncertain region or the most promising region in the surrogate model should be further evaluated, an exploration / exploitation trade-off. This is captured by the acquisition function, which determines which parameter configuration to evaluate next by optimizing the score assigned to the parameter configurations $x$ based on the predictions of the surrogate model. 
The point to be evaluated is expressed by $x_t = \argmin\limits_{x \epsilon \mathcal{X}} a(x)$, with an acquisition function $a$ and number of observations $t$. This expression is similar to the global optimum $x'$, the difference being that $a$ uses the inexpensive to evaluate surrogate model $g$, while $f$ is our expensive to evaluate objective function. 
With the candidate parameter configuration decided by the acquisition function, $f$ is used to obtain the $y_t$ observation: $y_t = f(x_t) + \phi_t$, with $\phi_t$ the noise in the observation at number of observations $t$. The optimal configuration can be extracted at any time: $x^+ = \argmin\limits_{x_i \epsilon \mathcal{X}_{1:t}} f(x_i)$, with $x_i$ the parameter combination evaluated at timestep $i$. 

Three acquisition functions are commonly used, and will be used here as well: Probability of Improvement, Expected Improvement and Upper Confidence Bound. 

Probability of improvement (\textit{PI}) is a conceptually simple acquisition function, which selects the next point based on the likelihood that it results in an improvement over the current optimum, $f(x^+)$~\cite{bayesian_optimization_pi_Kushner}. PI is expressed as $a_{PI}(x) = P(g(x) \leq (f(x^+) + \lambda))$, with $P$ the probability and a constant $\lambda \ge 0$ determining exploration / exploitation. 
Expected Improvement (\textit{EI}), in contrast to PI, includes discrimination on the size of the potential improvement~\cite{bayesian_optimization}. Expected improvement is expressed as $a_{EI}(x) = \mathbb{E} (\max(g(x) - f(x^+), 0))$, where $\mathbb{E}$ is the expected value. 
Another popular acquisition function, the Upper Confidence Bound (\textit{UCB}), is fundamentally different from PI and EI, as it is optimistic in the face of uncertainty, whereas PI and EI are based on improvement. 
UCB is very simplistic in the basis: $a_{UCB}(x) = \mu(x) + \lambda \sigma(x)$, where $\lambda$ is a non-negative parameter controlling exploration~\cite{BO_review}. 

As we assume minimization, we use the minimization variants of these acquisition functions. The minimization variant of UCB is known as Lower Confidence Bound (\textit{LCB}).  

\subsection{General structure}\label{sec:design-implementation-general-structure}
The GPU auto-tuning problem domain poses two main challenges for application of BO: the representations of parameter types and dealing with invalid configurations. 

\subsubsection{Parameter types}
First, parameters can be of many types: integers, real numbers, strings, Booleans, etc. We must thus assume that there can be a mixture of parameter types. In contrast, most work on BO assumes continuous parameters in the form of real numbers.
Traditionally, the solution to this problem has been to treat the search space as continuous and snap the continuous values to discrete values before evaluation. As described by~\citet{BO_categorical_integer}, an issue with this approach is the possible mismatch between the optimum given by the acquisition function and the actual set of parameter configurations, which can result in an inaccurate surrogate model and getting stuck on a single parameter configuration. 

Solving these kinds of problems has recently received attention. 
~\citet{BO_categorical_integer} propose a modified covariance function for integer and one-hot mapped categorical parameters, where the covariance is zero for impossible parameter values, effectively removing the distance between the discrete values. 
Similarly,~\citet{discrete_variable_BO} propose manipulation of the exploration factor of the acquisition function and the length scale of the covariance function to prevent the acquisition function from getting stuck. 
~\citet{mixed_BO} propose a mixed-variable BO approach named \textit{MiVaBO}, using a linear surrogate model and Thompson sampling. 
Likewise,~\citet{mixed_discrete_cont_variables_surrogate_models} propose several surrogate models for mixed discrete-continuous variables. 
~\citet{discrete_BO} observe that optimization of the acquisition function is more challenging for discrete search spaces, as gradient-based optimization can not be applied. They propose \textit{Amortized Bayesian Optimization}, which is the usage of a generative model that can be used longer than a single acquisition function optimization. 
~\citet{BO_continuous_categorical_CoCaBO} propose a framework for multiple continuous and categorical inputs named \textit{CoCaBO}, where multiple-armed bandits are used to select categorical inputs in addition to GP-based BO. 

Non-linearity is another problem surrounding parameter values, as in many computer-related parameters are non-linear, such as powers of two. 
Given that these parameters can not be assumed to refer to spatial units such as with kriging, this leads to a distortion within the surrogate model due to the distances between values. To avoid this, we normalize all numerical inputs in a linear fashion, mapping them back to the original value before outputting. 

Categorical values also pose a challenge when encoding. 
Simple integer encoding imposes an ordering on categorical values. Within machine learning, it is thus common practice to instead apply one-hot encoding in such situations, which would involve adding a binary dimension for every categorical value. Due to BO's unsuitability for high-dimensional problems, this is not an ideal solution. 
Instead, binary encoding could be a viable solution, given that it is much more efficient on the number of dimensions used compared to one-hot encoding, and does not suffer from the assumptions of integer encoding. 
However, a direct comparison does not seem to be in favor of binary encoding~\cite{encoding_binary_vs_onehot_vs_hasing}. 
Due to the structure of Kernel Tuner inputs, binary encoding has not been applied, as the user is responsible for the ordering of the parameters. 

Given the problems mentioned, it might seem odd to use a continuous surrogate model in the first place. However, as seen in \cref{sec:gaussian-processes}, discrete surrogate models have their drawbacks as well, and as shown by~\citet{discrete_BO_using_continuous}, continuous surrogate models can be well-suited for optimization of discrete problems.

\subsubsection{Invalid configurations}
The second challenge is that of the invalid parameter configurations, creating points in the search space that should not be selected. 

For GPU kernels, a configuration can be diagnosed to be invalid in three stages of the process: by checking the validity of individual parameters according to the programming model specification beforehand, via an error at compile time, and via an error at runtime. 
The latter can for example be due to an invalid configuration of thread blocks that are within programming model specification, but invalid on the actual GPU due to hardware restrictions, which is part of what makes this application specifically geared towards GPU kernels. 
The earlier invalid configurations can be detected, the better, as this avoids waste of time and computing resources for the user. To aid in this, Kernel Tuner allows users to impose restrictions on search spaces. However, we may not be able to detect all invalid configurations beforehand, and these thus need to be properly handled. 
This is more challenging than it may seem at first glance. If we do not fit an observation for the invalid parameter configuration to the surrogate model, the same invalid parameter configuration will always be chosen by the acquisition function.

Setting an observation value of our choice (given that an invalid parameter configuration can not have an observation value), is not ideal either. 
For example, a simple solution might seem be setting the observed value for such invalid configuration in the surrogate model to the inverse of the absolute optimum (an uninformative absolute optimum being $-\infty$ when minimizing and $\infty$ when maximizing), so the acquisition function will not choose this parameter configuration again. 
Alternatively, the observed value could be set to the mean of all observations so far. As variance is an important factor in acquisition functions that are not purely exploiting, the invalid parameter configuration is unlikely to be chosen again by the acquisition function. Despite the fact that no information has been gained due to the invalid parameter configuration other than that it is invalid, both options distort the surrogate model in such a way that the acquisition function either gets stuck or is less likely to choose parameter configurations near an invalid parameter configuration. While it is possible that neighbouring configurations are also invalid, it is impossible to know in advance exactly which parameter configurations will be invalid, so affecting the surrogate model with an artificial observation value is undesirable.

To address both the parameter type and invalid configuration problems, we propose the following structure for the application of BO to the problem domain: a discrete, normalized search space, where the acquisition function is solely optimized on the non-evaluated parameter configurations. 
This avoids spatial representation issues and prevents the acquisition function from getting stuck. 
In addition, it allows invalid configurations to be ignored, without distorting the surrogate model to achieve this. 
As Kernel Tuner already reports an average over a user-defined number of runs per evaluation, there is little practical need to revisit evaluated parameter configurations.

\subsection{Initial sampling}\label{sec:design-implementation-initial-sampling}
Given the black-box nature of BO, it is possible to start optimization on a completely unknown surrogate model. 
However, it is common to first take an initial sample of the search space to aid exploration. 
While random sampling is often used, it is likely that the samples are unevenly distributed throughout the search space. 
As such,~\citet{Miyazaki_HPC_BO} suggest the use of Latin Hypercube Sampling (LHS). This technique ensures that the initial samples are spread evenly throughout the search space, providing a more balanced initial sample. 

In practice, this can be made difficult due to the possibility of invalid configurations. 
As we want to explore large search space in as few evaluations as possible, a small Latin Hypercube Sample of a large search space is in principle more balanced than random sampling, but also more easily unbalanced when one or more samples are missing due to invalidity. 
To avoid this, we thus first apply Latin Hypercube Sampling. If there are invalid samples, these are replaced by random samples until all initial samples are valid. 
This combination avoids a skewed initial sample.

\subsection{Dynamic exploration factor}\label{sec:design-implementation-dynamic-exploration-factor}
As seen in \cref{sec:acquisition-functions}, the acquisition functions have an exploration factor, which is often set to a constant value. 
However, it would intuitively seem more sensible to make an informed calculation for this hyperparameter based on the state of the surrogate model at every evaluation. 
This is what~\citet{contextual_improvement_explore_exploit} proposed with \textit{contextual improvement}, a variation on the probability of improvement using \textit{contextual variance} for the $\lambda$ exploration parameter, defined as $\frac{\overline{\sigma^2}}{f(x^+)}$, where $\overline{\sigma^2}$ is the mean of variances in the posterior surrogate model and $f(x^+)$ is the best found observation so far. 
The experiments by~\citet{practical_BO_contextual_variance} suggest that the constant $\lambda=0.01$ works well in most cases, while counter-intuitively, setting a cooling schedule for the exploration factor does not work as well. This could mean contextual variance does not work well either. 
However,~\citet{contextual_improvement_explore_exploit} demonstrate that their contextual improvement can outperform a constant factor, and in addition avoids dependence on this parameter. Hence, such an approach to exploration / exploitation can be of benefit given that we want to be our implementation to be as generally applicable as possible. 

However, the contextual variance can not be applied as-is to our problem domain: it is intended for maximization, and behaves differently depending on the absolute scale of the observations. The latter means that given two maximization problems with a similar mean variance, for a problem with a lower maximum, exploration will be disproportionately large, whereas for a problem with a higher maximum, exploration will be practically ignored. This is undesirable behaviour, and as such, applying this contextual variance is not simply a matter of setting the function to the inverse to obtain the minimum, but requires a new function instead. 

We propose a function that outputs a hyperparameter $0 \le \lambda$, which scales in proportion to $f(x^+)$, the mean variance, and the initial sample mean. 
To overcome the observation size dependence, we use the initial sample mean $\mu_{s}$ to get the fraction of improvement of the initial sample mean to the current optimal. 
To normalize the hyperparameter, we assume $0$ as the minimum and the mean of variances in the posterior surrogate model after the initial sampling, $\overline{\sigma_s^2}$, as the maximum. 
This results in the following function: $\lambda = \Big( \frac{\overline{\sigma^2}}{ \mu_{s} / f(x^+) } \Big) \big/ \overline{\sigma_s^2}$.

\subsection{Implementation}\label{sec:design-implementation-implementation}

Several Python packages already implement BO, including \textit{Bayesian Optimization}~\cite{BayesianOptimization_PythonPackage}, \textit{SciKit-Opt}~\cite{scikit-optimize} and \textit{HyperOpt}~\cite{hyperopt_tpe}. However, we need to customize our BO implementation to the problem domain and therefore we create our own implementation on top of SciKit-learn's \textit{Gaussian Process Regressor}~\cite{scikit-learn}.

While the existing implementations use techniques such as BFGS to optimize the acquisition function, due to the discrete structure of our search spaces this is not practical for our implementation. Instead, we exhaustively predict every discrete point in the model. To compensate, we reuse the predictions when using multiple acquisition functions.

We propose a '\textit{multi}' acquisition function, which skips acquisition functions when they repeatedly suggest the same candidates. 
We register how many duplicates were suggested per acquisition function in the past iterations. When this count exceeds a threshold, the \textit{skip threshold}, the conflicting acquisition functions are pitted against each other. 
The acquisition function with the lowest discounted observations score is kept, while the others will be skipped, encouraging efficiency. 
This discounted observation score is calculated per acquisition function by ${dos_t} = \sum_{i=1}^{t} o_i \cdot 0.9^{t-i}$, where $t$ is the current iteration number, $o$ is the vector of observations over time, and $0.9$ is the \textit{discount factor}. This gives more recent observations more weight. 
As the surrogate model becomes more accurate and the acquisition functions suggest the same candidates more often, gradually fewer acquisition functions will be used. Eventually a single acquisition function will remain, returning to the traditional approach of a single acquisition function. 

In addition, we propose an '\textit{advanced multi}' acquisition function similar to the '\textit{multi}' acquisition function, but removing predictions to avoid duplicates and directly judging acquisition functions based on their discounted observations score instead. Invalid observations use the median of the valid observations in this score, to avoid skewing the score. 
If an acquisition function scores more than 5\% above the mean of the discounted observations (assuming minimization) for \textit{skip threshold} or more times, this acquisition function will be skipped and the counts of others reset. 
If an acquisition function scores more than 5\% below the mean of the discounted observations for \textit{skip threshold} or more times, this acquisition function will be promoted to be the only acquisition function for the remainder of the optimization. We will call this threshold of 5\% in both cases the \textit{required improvement factor}. 
This approach is intended to progressively select the optimal acquisition function for a specific problem. 

The acquisition functions we propose are different from \textit{GP-Hedge}~\citet{acquisition_functions_multiple_portfolio}, which attempts to select the optimal acquisition function per function evaluation, requiring full prediction and optimization of all acquisition functions at every function evaluation. 
Our methods require fewer predictions, evaluating acquisition functions in a round-robin fashion, optimizing one acquisition function per function evaluation, until the best-performing acquisition function remains. 

\subsection{Hyperparameter tuning}\label{sec:optimization}

To further optimize the implementation and provide valuable defaults to Kernel Tuner users, we tune the hyperparameters of the initial sampling, surrogate model, and acquisition functions. 
The hyperparameters were tuned on the kernels described in \cref{tab:kernels-specification} (see \cref{sec:methods}) on the Nvidia GTX Titan X GPU. 
The resulting optimal hyperparameters are displayed in \cref{tab:hyperparam_optimizations}. 
\begin{table}[htbp]
\centering
\small
\resizebox{\linewidth}{!}{%
\begin{tabular}{ll}
\textbf{Hyperparameter}              & \textbf{Best value(s)}              \\ \hline
Covariance function lengthscale      & 3/2 2                               \\
Exploration factor                   & CV                                  \\
Covariance function lengthscale (CV) & 3/2 1.5                             \\
Skip threshold                       & 5                                   \\
Order of acquisition functions       & (ei, poi, lcb)                      \\
Required improvement factor          & 0.1                                 \\
Discount factor                      & 0.65 (multi), 0.75 (advanced multi) \\
Initial sampling                     & maximin                             \\
Pruning                              & yes                                 \\
Acquisition functions                & advanced multi, multi, EI           \\ \hline
\end{tabular}%
}
\caption{Result of hyperparameter optimization}
\label{tab:hyperparam_optimizations}
\end{table}

\section{Experimental results}\label{sec:results}

In this section, we evaluate the performance of our BO implementation for auto-tuning GPU kernels. \cref{sec:methods} describes the experimental setup. \cref{sec:results-comparison} compares our methods to the other Kernel Tuner methods, and \cref{sec:results-comparison-other-frameworks} compares our methods to other BO frameworks. 
In addition, \cref{sec:results-comparison-multidevice,sec:results-comparison-new-kernel} demonstrate how our methods generalize across GPUs and kernels respectively.

\subsection{Methods}\label{sec:methods}

The search space depends on the tunable kernels as well as the hardware, hence we employ various GPUs to evaluate our search strategies. 
First is the NVIDIA GTX Titan X with the \textit{Maxwell} architecture, which contains 3072 cores in 24 streaming multiprocessors with 12 GB GDDR5 memory~\cite{nvidia_gtx_titan_x}. We also evaluate the performance of our BO implementation on the NVIDIA RTX 2070 Super (2019, \textit{Turing} architecture, 2560 cores, 40 SMs, 8 GB GDDR6 memory~\cite{nvidia_rtx_2070_super}) and the NVIDIA A100 (2020, \textit{Ampere} architecture, 6912 cores, 108 SMs, 40 GB HBM2e memory~\cite{nvidia_A100_40GB}) in \cref{sec:results-comparison-multidevice}. 
 
In this evaluation, we use three highly-tunable GPU kernels to evaluate and compare our optimization strategies, namely an OpenCL general dense matrix multiplication (\textit{GEMM}) kernel, a \textit{2D Convolution} kernel implemented in CUDA and a heterogeneous point-in-polygon (\textit{PnPoly}) kernel. Their variation in programming model, type of calculation and tunable parameters are intended to serve as a diverse sample of GPU kernels. 

\begin{table}[tbp]
{\centering
\setlength\tabcolsep{1pt}
\scriptsize
\begin{tabular}{llll>{\raggedright\arraybackslash}p{3.3cm}<{}}
\toprule
\textbf{Kernel} & \textbf{Configurations} & \textbf{Invalid} & \textbf{Minimum} & \textbf{Tunable parameters}      \\
\midrule
GEMM            & 17956                      & 0 (0\%)          & 28.307        & $M_{wg}$, $N_{wg}$, $K_{wg}$, $M_{dimC}$, $N_{dimC}$, $M_{dimA}$, $N_{dimB}$, $K_{WI}$, $V_{WM}$, $V_{WN}$, $S_{TRM}$, $S_{TRN}$, $S_A$, $S_B$, PRECISION \\
\midrule
Convolution     & 9400                       & 3624 (38.5\%)    & 1.625         & filter\_width, filter\_height, block\_size\_x, block\_size\_y, tile\_size\_x, tile\_size\_y, use\_padding, read\_only           \\
\midrule
PnPoly          & 8184                       & 323 (3.9\%)      & 26.968          & block\_size\_x, tile\_size, between\_method, use\_precomputed\_slopes, use\_method \\
\bottomrule
\end{tabular}%
\caption{Specifications of tunable kernels for the GTX Titan X. Minimum execution time is given in milliseconds.}
\label{tab:kernels-specification}
}
\end{table}

\begin{itemize}
\item \textbf{GEMM.} The GEMM kernel is based on the tunable OpenCL GEMM kernel found in CLBlast~\cite{CLBlast}. GEMM has a large number of tunable parameters (15 in total), which describe the thread block dimensions, the square area on which each thread block operates, whether or not to use shared memory for either input matrix, and vector widths. The Cartesian product of these parameters has a size of 82944, the search space has a size of 17956 after application of restrictions, giving GEMM the largest constrained search space of all test kernels.

\item \textbf{Convolution.} The final GPU kernel is a 2D Convolution kernel implemented in CUDA, specifically one for image filtering, an extension of the 2D convolution kernel presented in~\cite{convolution}. The tunable parameters describe the thread block dimensions, work per thread, and whether or not to use padding in shared memory to avoid shared memory bank conflicts and whether or not to use the read-only cache. The Cartesian product of these parameters has a size of 18432, after application of restrictions the search space contains 9400 parameter configurations. 

\item \textbf{PnPoly.} The Point-in-Polygon (PnPoly) kernel is taken from~\cite{goncalves2016spatial}. This is a heterogeneous kernel, where most of the computation happens on the GPU, but part of the computation calculation can be performed on the CPU (depending on a tunable parameter). The kernel overlaps CPU-GPU data transfers with computations on the GPU. As such, the CPU-GPU data transfers are also included in the run time. The other tunable parameters describe the thread block dimensions, work per thread, and switches between different algorithms for certain calculations. PnPoly has no restrictions applied, so the search space is a Cartesian product totalling 8184 configurations. 

\end{itemize}

As a baseline for our experiments, every run of our BO search strategies uses 20 function evaluations for taking an initial sample and 200 function evaluations for the optimization process itself. Plots set off to the number of function evaluations will thus start at 20 and end at 220. 
Each run of the BO methods is repeated 35 times. The other search methods in Kernel Tuner and other frameworks used for comparison are repeated 35 times as well, while the random sample method is repeated 100 times due to the greater variance. 
For all plots lower is better unless otherwise noted. 

To compare performance of an optimization methods over multiple kernels, we use the mean of the factor of deviation from the kernel mean absolute error mean, designated with \textit{Mean Deviation Factor} (\textit{MDF}). 
The mean absolute error (\textit{MAE}) is calculated against the global minimum of the search space over the function evaluations from 40 to 220, as the first evaluations are relatively noisy and their results depend too much on the quality of the initial sample. More specifically, we calculate the mean absolute error by $\frac{1}{10} \sum_{i=2}^{11} | f(x_{20i}^+) - f(x') |$, with $x'$ the globally optimal parameter configuration and $x_{20i}^+$ the best found parameter configuration after $20i$ function evaluations. The mean MAE across the runs is used for the MDF. 
The mean deviation factor is used in the form of a bar chart, where the error bars show standard deviation of the factors.  
This MDF is calculated by taking the mean MAE for each kernel, dividing it by the mean of the mean MAE of all kernels. This quantifies the performance of a strategy, allowing for comparison of approaches across multiple kernels, while retaining differences in MAE without the influence of varying performance scales between kernels.

\subsection{Comparison}\label{sec:results-comparison}

\begin{figure*}[p]
  \begin{minipage}[t]{0.33\textwidth}%
    \centering
    \subcaptionbox{GEMM \label{fig:results_comparison_GEMM}}{\includegraphics[width=\linewidth]{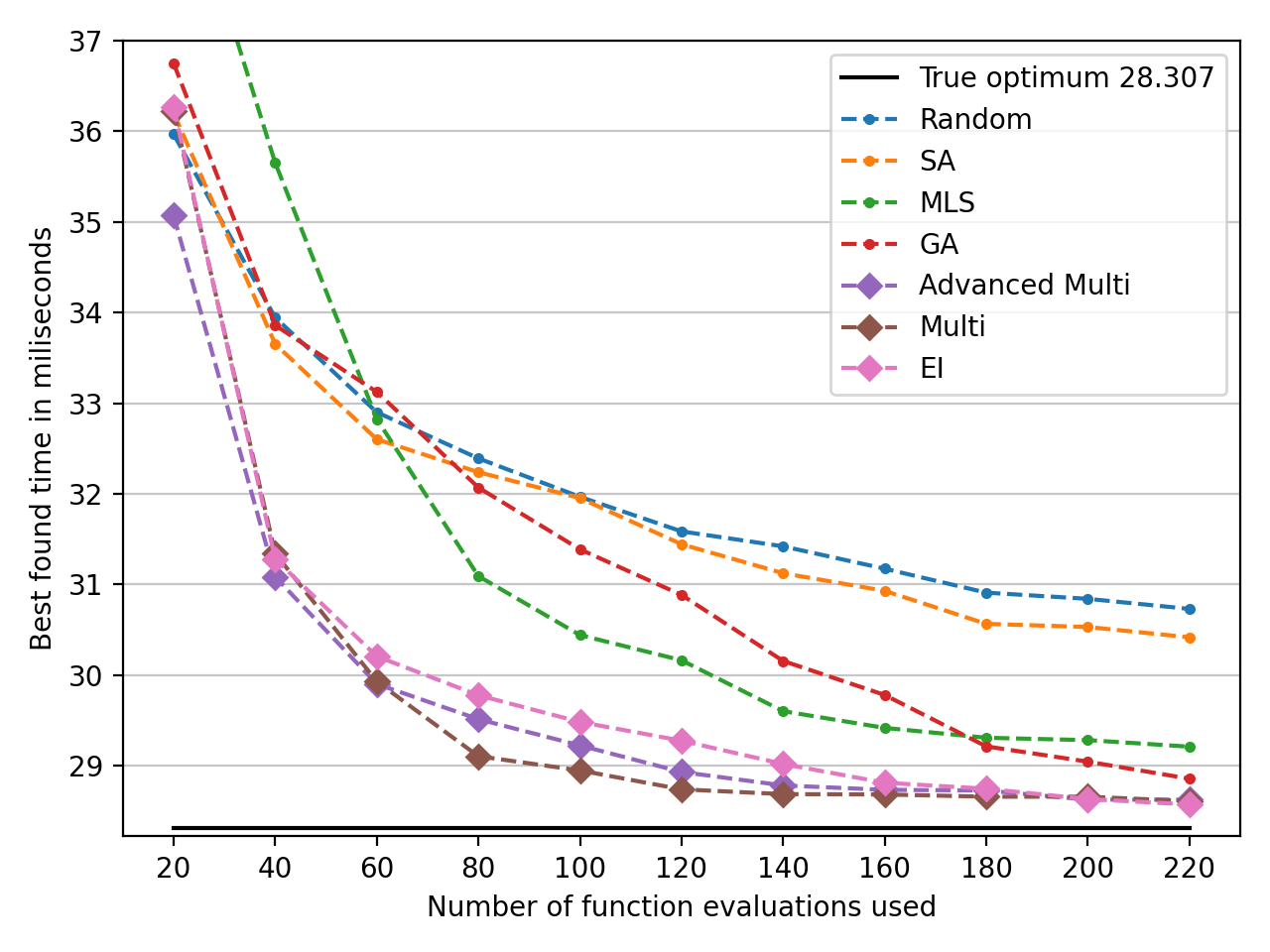}}
    \subcaptionbox{Convolution \label{fig:results_comparison_convolution}}{\includegraphics[width=\linewidth]{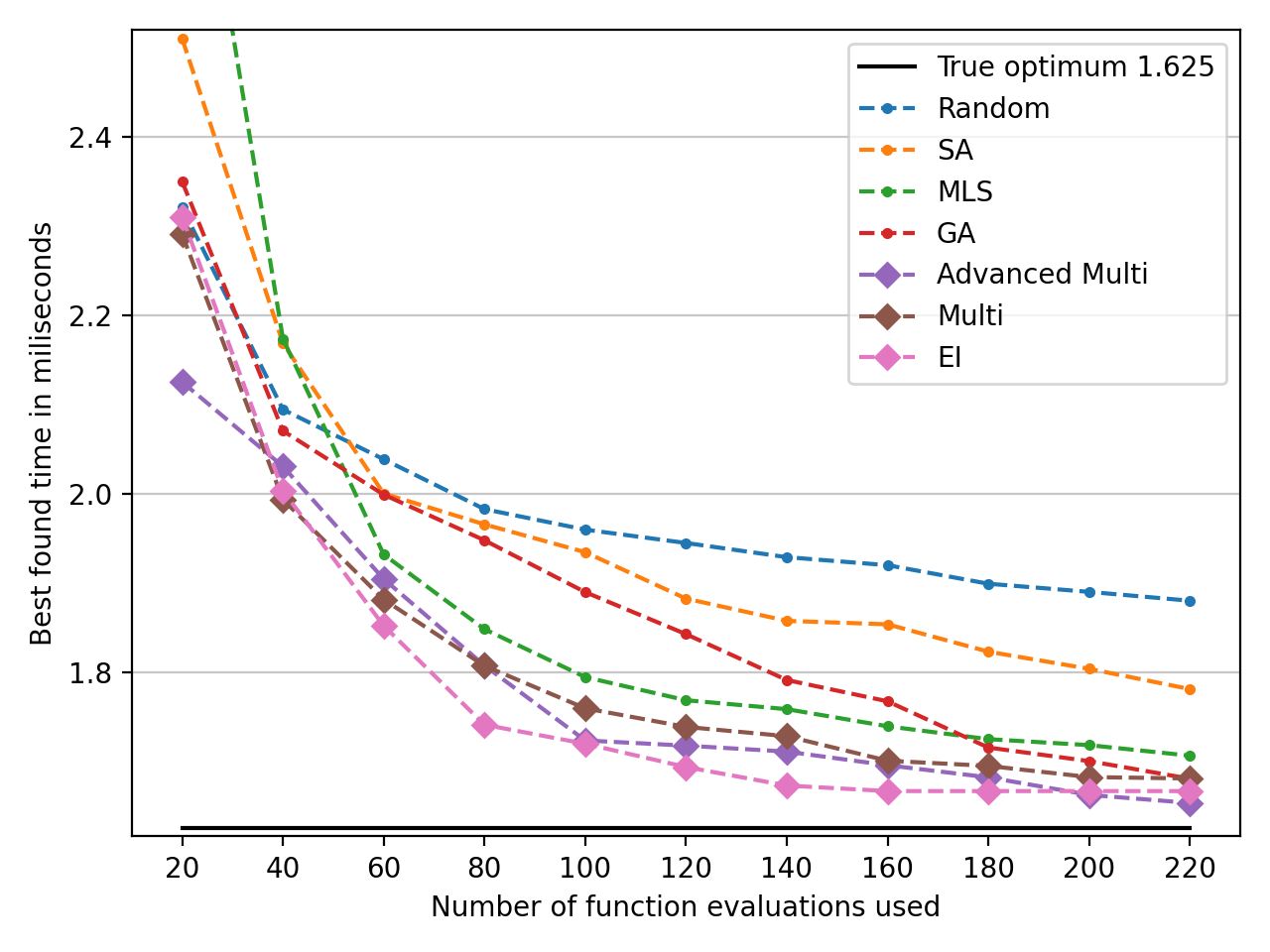}}
    \subcaptionbox{PnPoly \label{fig:results_comparison_pnpoly}}{\includegraphics[width=\linewidth]{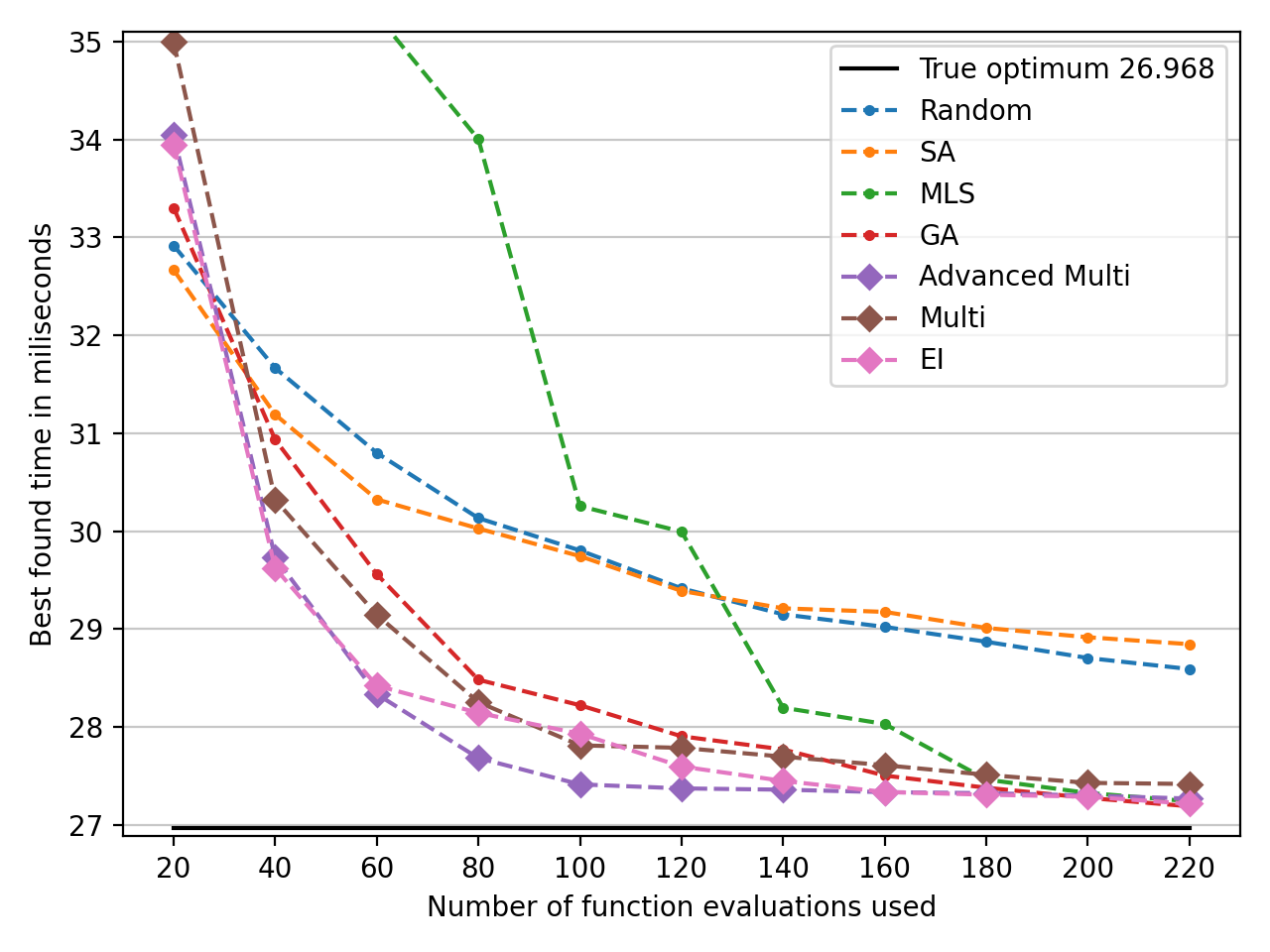}}
    \subcaptionbox{Mean deviation factors \label{fig:results_comparison_all_strategies_RMSE_GPU}}{\includegraphics[width=\linewidth]{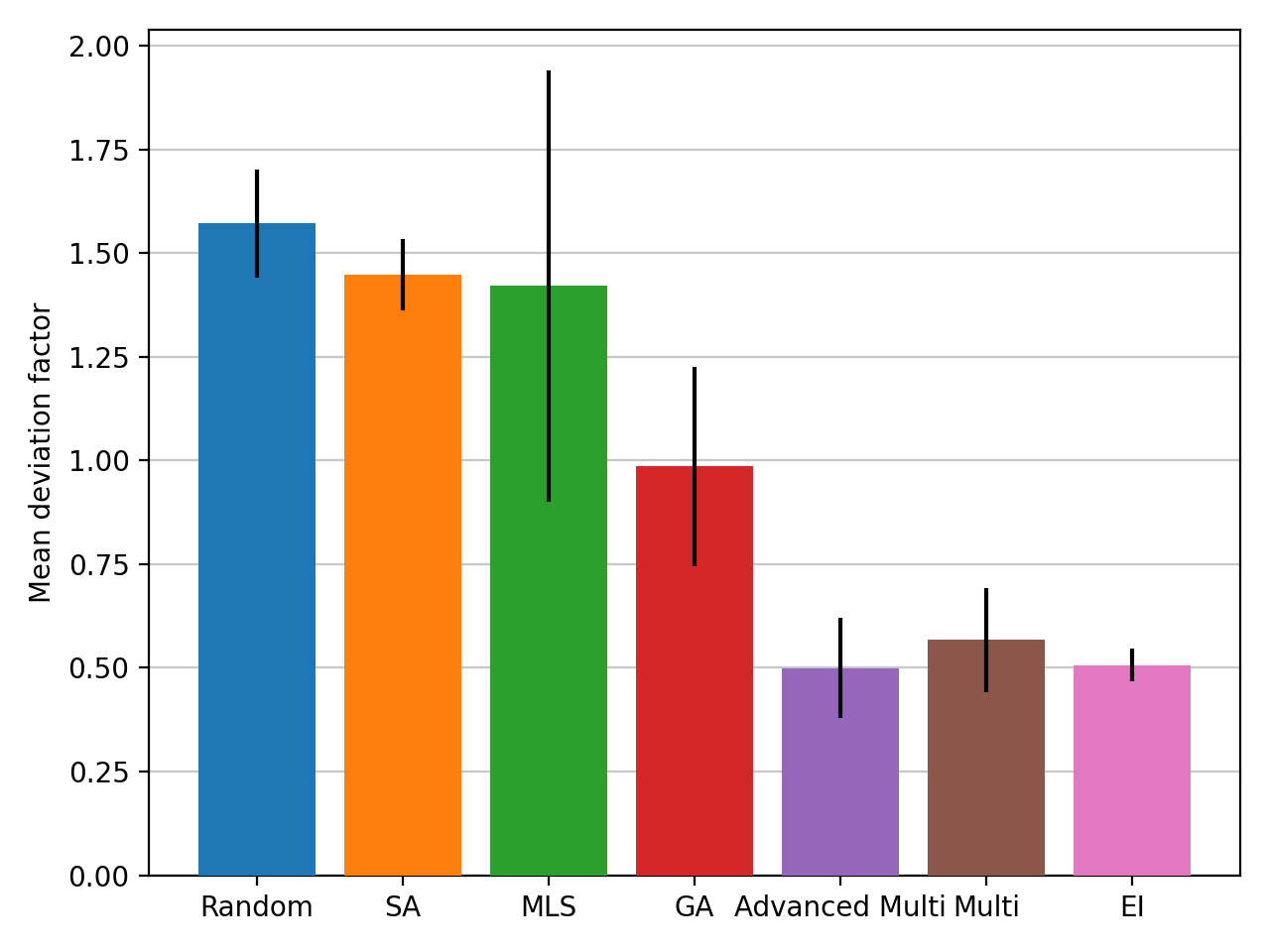}}
    \caption{Comparison of search methods on GPU kernels with the GTX Titan X; \cref{fig:results_comparison_GEMM,fig:results_comparison_convolution,fig:results_comparison_pnpoly} show how the best found performance per method increases as the number of function evaluations increases.}
    \label{fig:results_comparison_searchmethods}
  \end{minipage}%
  \quad
  \begin{minipage}[t]{0.33\textwidth}%
    \centering
    \subcaptionbox{GEMM \label{fig:results_comparison_gpu_rtx_2070_super_GEMM}}{\includegraphics[width=\linewidth]{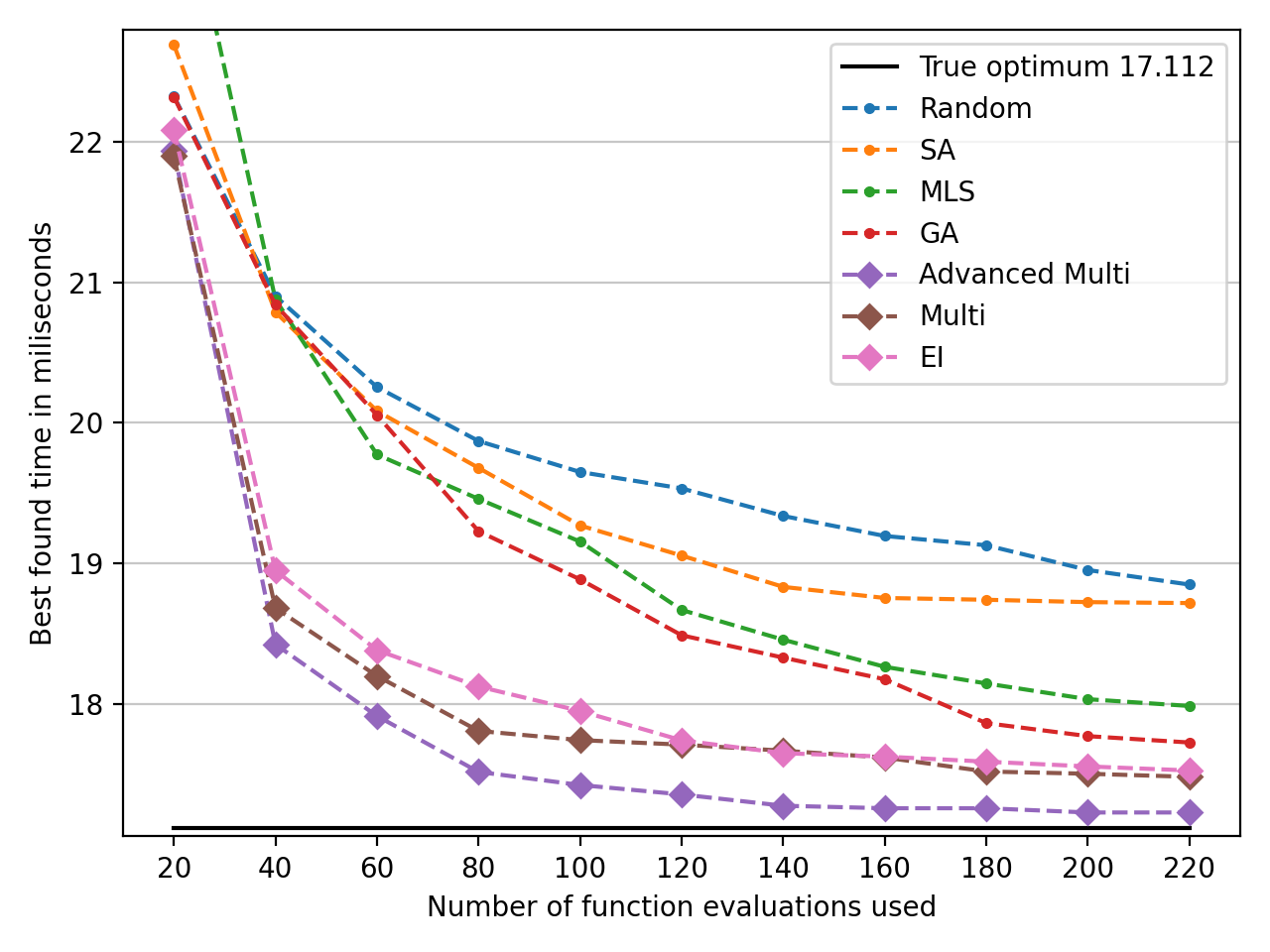}}
    \subcaptionbox{Convolution \label{fig:results_comparison_gpu_rtx_2070_super_convolution}}{\includegraphics[width=\linewidth]{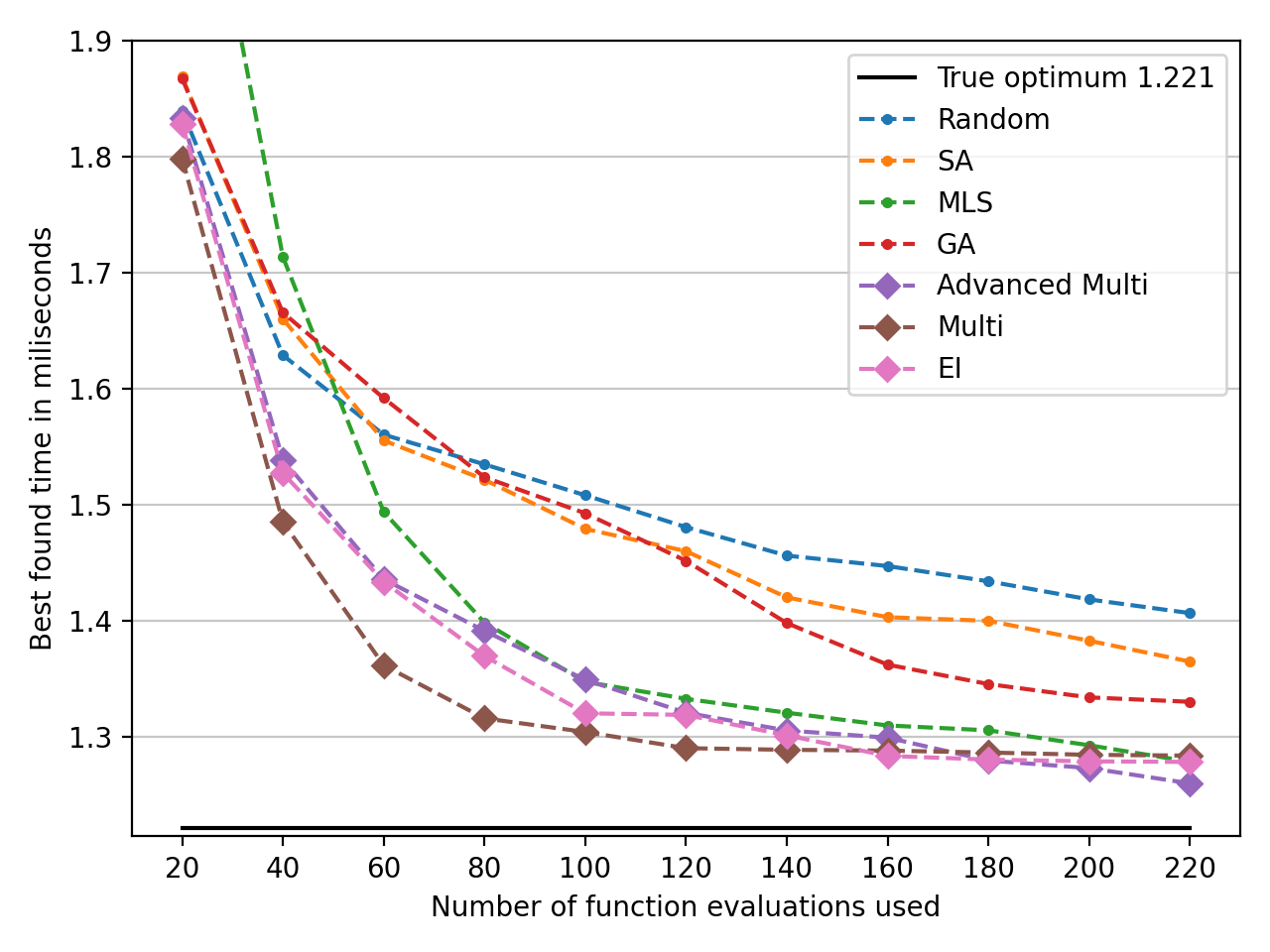}}
    \subcaptionbox{PnPoly \label{fig:results_comparison_gpu_rtx_2070_super_pnpoly}}{\includegraphics[width=\linewidth]{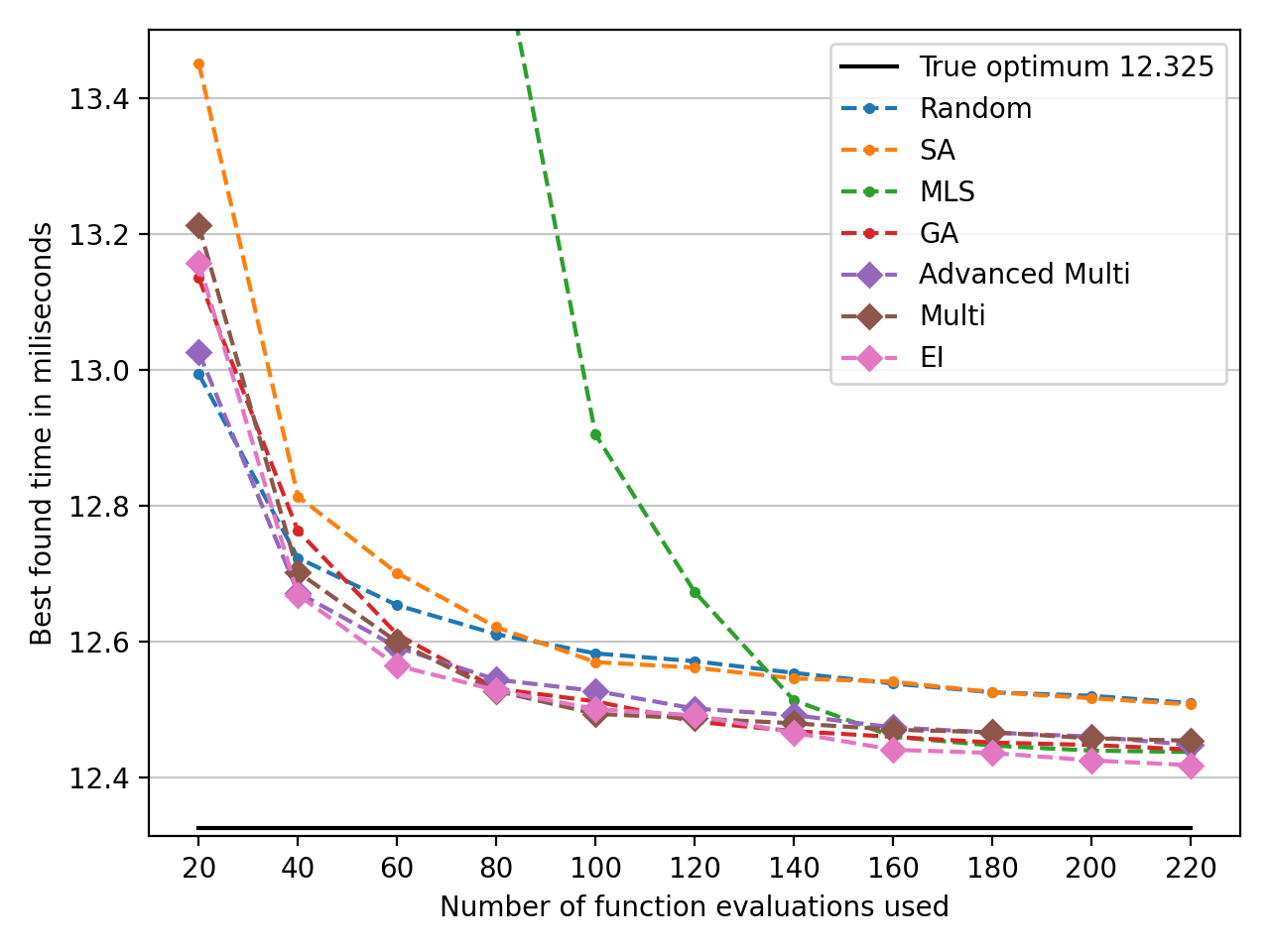}}
    \subcaptionbox{Mean deviation factors \label{fig:results_comparison_gpu_rtx_2070_super_factors}}{\includegraphics[width=\linewidth]{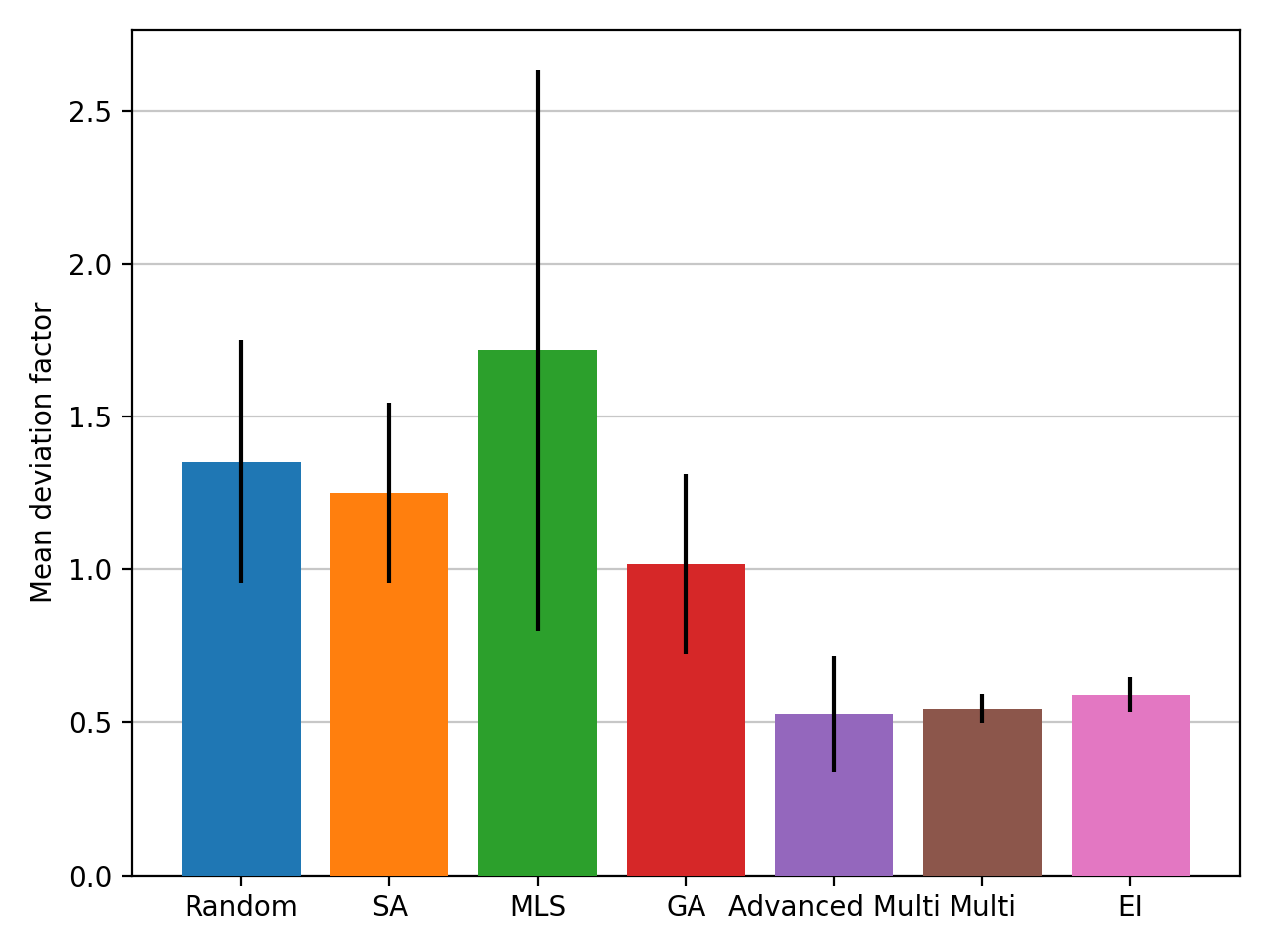}}
    \caption{Comparison of search methods on GPU kernels with the RTX 2070 Super; \cref{fig:results_comparison_gpu_rtx_2070_super_GEMM,fig:results_comparison_gpu_rtx_2070_super_convolution,fig:results_comparison_gpu_rtx_2070_super_pnpoly} show how the best found performance per method increases as the number of function evaluations increases.}
    \label{fig:results_comparison_gpu_rtx_2070_super}
  \end{minipage}%
  \quad
  \begin{minipage}[t]{0.33\textwidth}%
    \centering
    \subcaptionbox{GEMM \label{fig:results_comparison_gpu_A100_GEMM}}{\includegraphics[width=\linewidth]{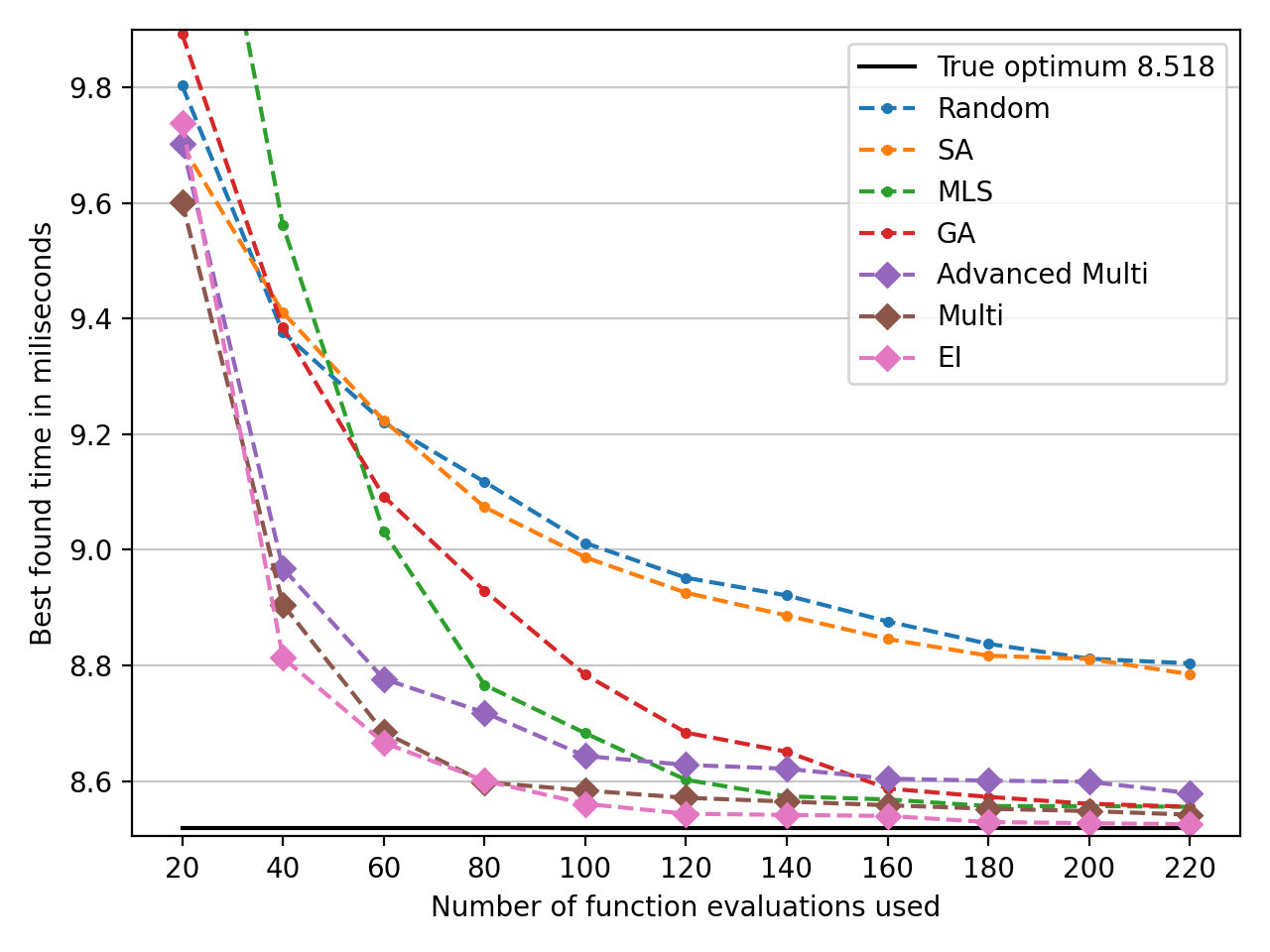}}
    \subcaptionbox{Convolution \label{fig:results_comparison_gpu_A100_convolution}}{\includegraphics[width=\linewidth]{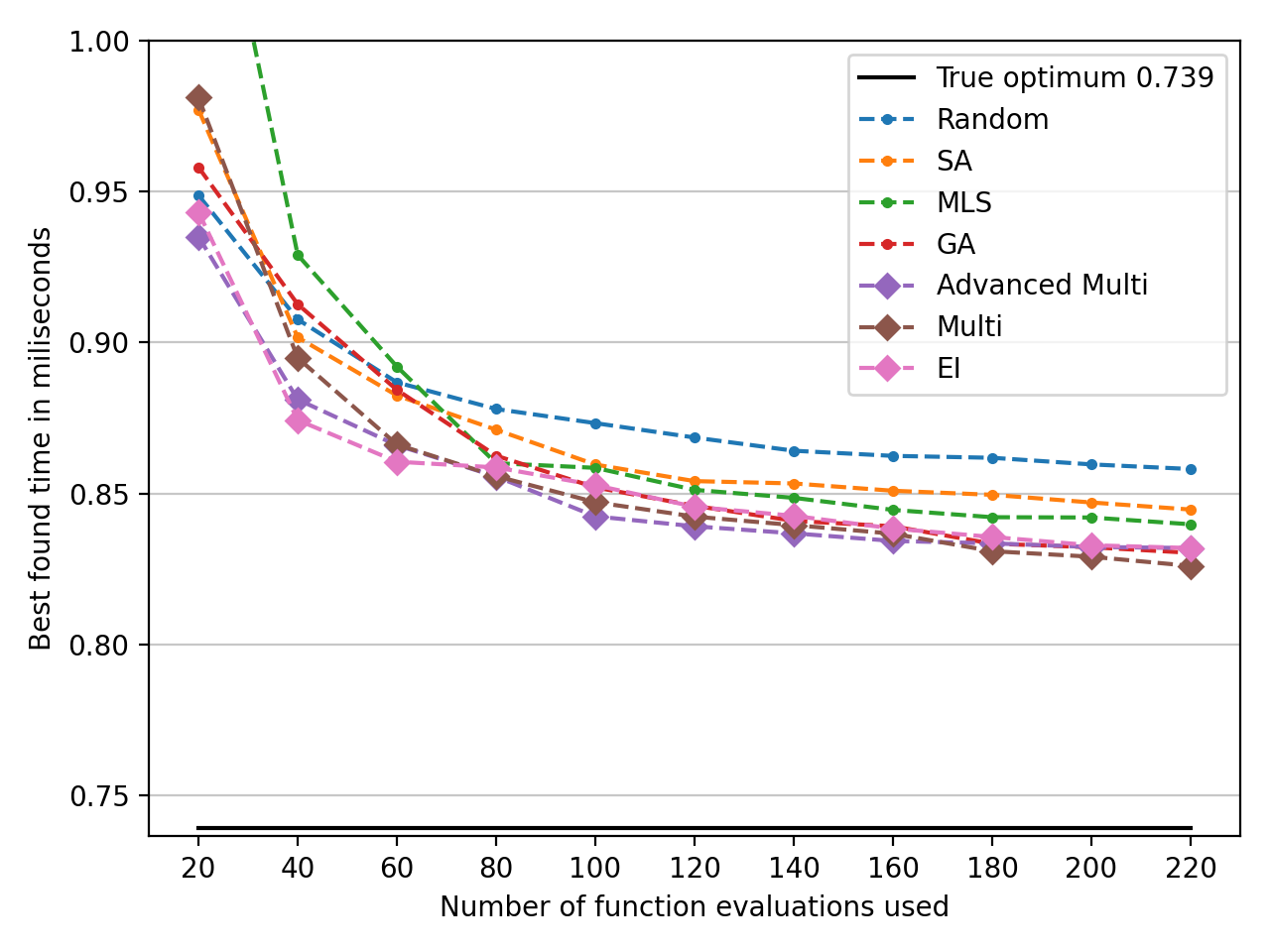}}
    \subcaptionbox{PnPoly \label{fig:results_comparison_gpu_A100_pnpoly}}{\includegraphics[width=\linewidth]{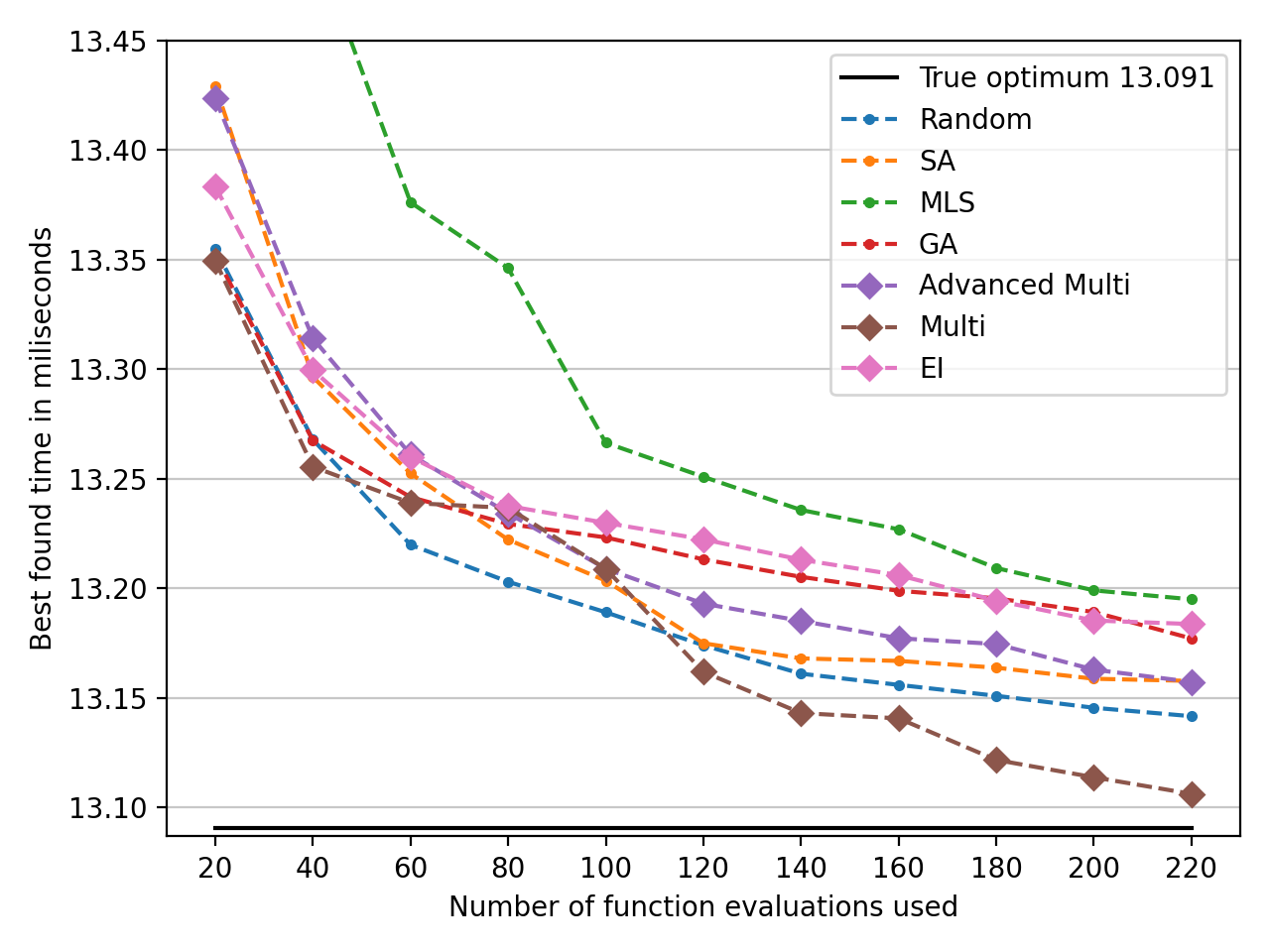}}
    \subcaptionbox{Mean deviation factors \label{fig:results_comparison_gpu_A100_factors}}{\includegraphics[width=\linewidth]{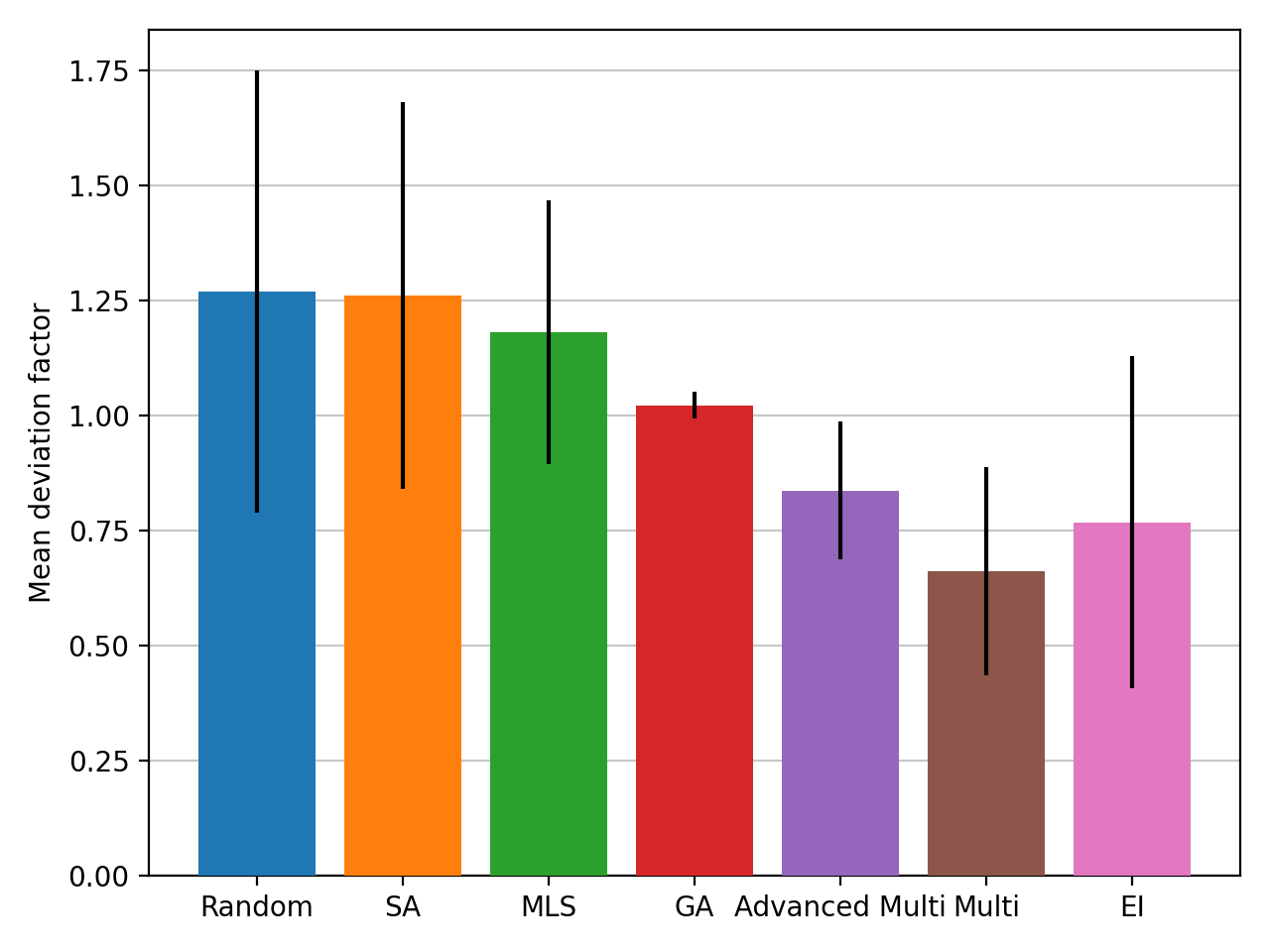}}
    \caption{Comparison of search methods on GPU kernels with the A100; \cref{fig:results_comparison_gpu_A100_GEMM,fig:results_comparison_gpu_A100_convolution,fig:results_comparison_gpu_A100_pnpoly} show how the best found performance per method increases as the number of function evaluations increases.}
    \label{fig:results_comparison_gpu_A100}
  \end{minipage}%
\end{figure*}

We compare the performance of our Bayesian Optimization method for a number of selected acquisition functions with the other optimization methods in Kernel Tuner, as well as through an extended comparison based on how long it takes other search methods to achieve performance similar to our search strategies. 
The search methods in Kernel Tuner selected for comparison are Simulated Annealing (\textit{SA}), Multi-start Local Search (\textit{MLS}) and Genetic Algorithm (\textit{GA}), as these are the methods that performed best on the test kernels. 
The diamonds mark our methods, while the dots mark the other methods in Kernel Tuner. 
For all three GPU kernels in \cref{fig:results_comparison_GEMM,fig:results_comparison_convolution,fig:results_comparison_pnpoly}, our methods appear to perform relatively similar.
Most importantly, our search methods consistently outperform the other Kernel Tuner search methods. 
MLS and GA perform relatively well in the end, especially GA, but as seen in \cref{fig:results_comparison_all_strategies_RMSE_GPU}, our methods perform better overall. 
These results appear to be promising: our methods have a much lower mean deviation factor than the other Kernel Tuner methods. 

\begin{figure}[tbh]
    \centering
    \includegraphics[width=0.8\linewidth]{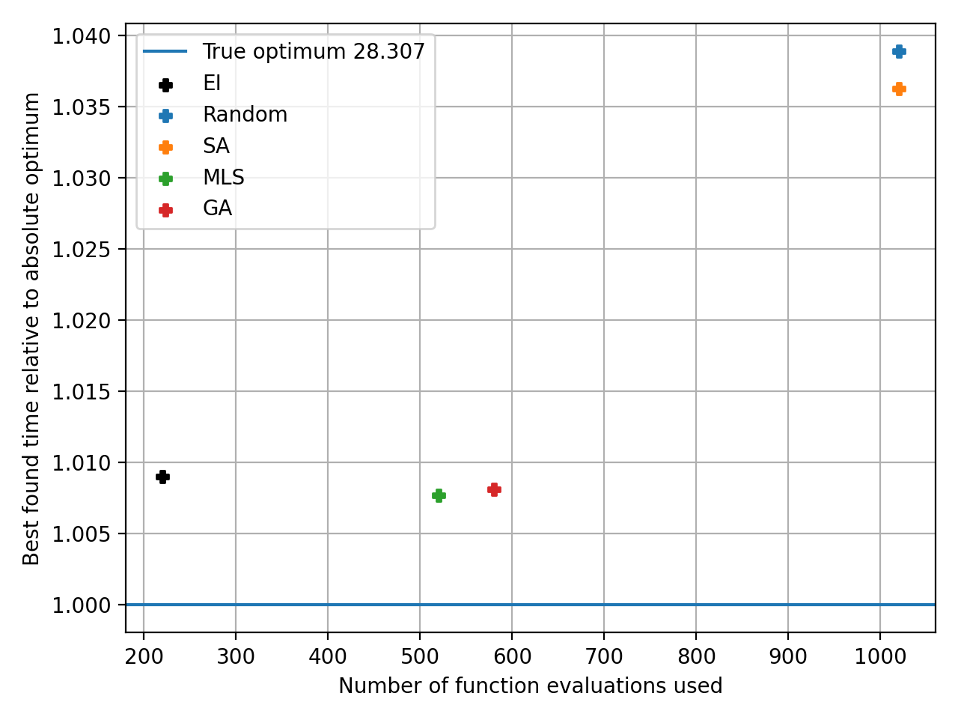}
    \caption{Extended relative performance on GEMM with the GTX Titan X}
    \label{fig:results_comparison_extended_GEMM}
\end{figure}

What has not been addressed is how hard or easy it is to find the optimum or near-optimum per search space; e.g. for some search spaces, it may be much harder to get to 95\% of the optimum than to get to 90\%. Due to space constraints it is not possible to show this for each combination of kernel and GPU, yet as an example we take GEMM on the GTX Titan X. 
\cref{fig:results_comparison_extended_GEMM} demonstrates the best found time of one of our acquisition functions (EI) at 220 function evaluations, and plots the other tuners at the number of function evaluations where they first match or exceed the best found time by EI, up to a maximum of 1020 function evaluations. 
In \cref{fig:results_comparison_GEMM} at 220 unique function evaluations, GA appears to be relatively close to our methods. Yet \cref{fig:results_comparison_extended_GEMM} shows that it takes GA nearly 3 times as many unique function evaluations to match the best found time of EI. MLS performs slightly better, while SA and random do not get close to matching the best found time in five times as many function evaluations.

\subsection{Other GPUs}\label{sec:results-comparison-multidevice}
At this point, we have exclusively used the Nvidia GTX Titan X GPU to compare our results. 
To see how our search strategies fare on more modern, different GPUs, we compare results on the RTX 2070 Super and the A100. 
While the tunable kernel parameters are kept the same, the change of GPUs means that this is essentially a different search space, with different sizes, minima, and invalid configurations. These details are shown in \cref{tab:GPU-kernel-details}. 

\begin{table}[tbp]
\centering
\resizebox{\linewidth}{!}{%
\begin{tabular}{lllll}
\hline
\textbf{GPU}                    & \textbf{Kernel} & \textbf{Search space size} & \textbf{Invalid} & \textbf{Minimum} \\ \hline
\multirow{3}{*}{RTX 2070 Super} & GEMM            & 17956                      & 0                & 17.112           \\
                                & Convolution     & 7520                       & 1744 (23.2\%)    & 1.221            \\
                                & PnPoly          & 8184                       & 288 (3.5\%)      & 12.325           \\ \midrule
\multirow{3}{*}{A100}           & GEMM            & 17956                      & 0                & 8.518            \\
                                & Convolution     & 7520                       & 1744 (23.2\%)    & 0.739            \\
                                & PnPoly          & 8184                       & 317 (3.9\%)      & 13.091           \\ \bottomrule
\end{tabular}%
}
\caption{Kernel details per GPU}
\label{tab:GPU-kernel-details}
\end{table}

The performance of the three GPU kernels on the RTX 2070 Super and A100 are shown in \cref{fig:results_comparison_gpu_rtx_2070_super,fig:results_comparison_gpu_A100} respectively. For GEMM, our search strategies perform comparable to the GTX Titan X in both cases. 
The performance on Convolution is good in both cases as well. It must be noted that for Convolution on the A100, the first two best configurations are at ~0.739 and ~0.761, while the third best is at 0.823, explaining the large difference between the true optimum and the obtained performance in \cref{fig:results_comparison_gpu_A100_convolution}. 
Also remarkable is the performance of the BO strategies for PnPoly on the A100, for which \textit{advanced multi} and EI perform worse than for PnPoly on the GTX Titan X and the RTX 2070 Super. However, when combined with the \textit{multi} acquisition function, BO is the only optimization method that is able to reliably return near-optimal configurations for PnPoly on the A100.
Looking at \cref{fig:results_comparison_gpu_rtx_2070_super_factors,fig:results_comparison_gpu_A100_factors}, we can see that our strategies generalize well to the kernels on other GPUs. The difference in the mean deviation factors between the A100 and GTX Titan X can mostly be explained by the relatively worse performance of \textit{advanced multi} and EI for PnPoly on the A100.

\subsection{Other frameworks}\label{sec:results-comparison-other-frameworks}

Following the comparison to the other methods in Kernel Tuner, we now compare our search strategies to alternative BO implementations. 
These are the widely-used \textit{BayesianOptimization} and \textit{Scikit-optimize} Python packages.  
We take the defaults recommended by the frameworks, as the hyperparameters used by us are optimized specifically towards our implementation. 
Neither of these frameworks is able to take search space constraints into account. 
\textit{BayesianOptimization} uses UCB with exploration factor $\kappa=2.576$ and 5 restarts for the acquisition function optimizer as the defaults. 
By default, Scikit-optimize uses \textit{GP-Hedge} with exploration factors $\xi=0.01$, $\kappa=1.96$ and 5 restarts for the acquisition function optimizer. 
We set the number of function evaluations and the number of initial samples to the same values as before. 
To create a fair comparison, we use the Nvidia RTX 2070 Super for the comparison among frameworks, as no framework has been optimized for the search spaces on this GPU specifically.

\begin{figure}[tbp]
    \centering
    \begin{subfigure}[b]{0.8\linewidth}
        \centering
        \includegraphics[width=\linewidth]{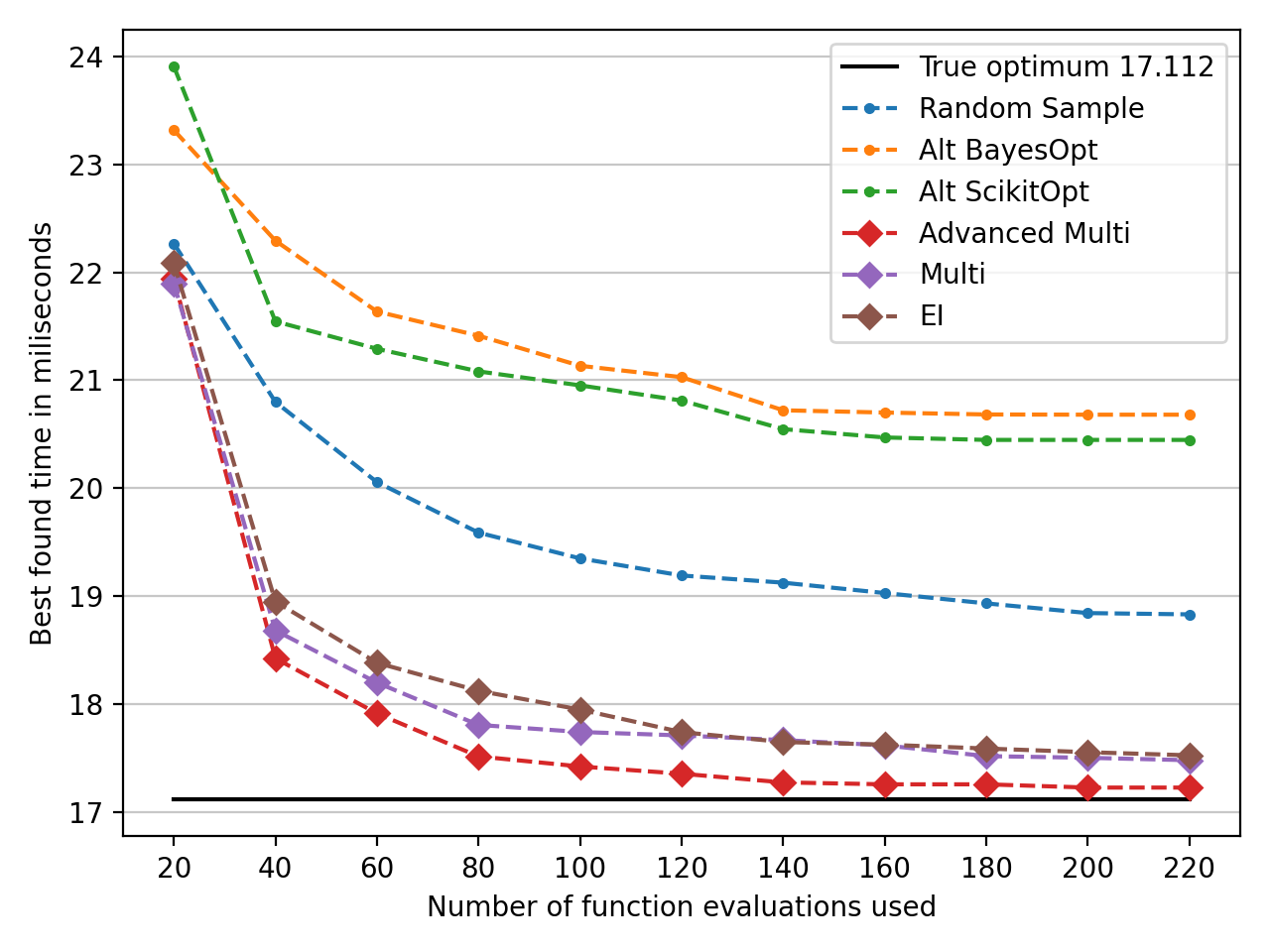}
        \caption{GEMM}
        \label{fig:results_frameworks_comparison_GEMM}
    \end{subfigure}
    \begin{subfigure}[b]{0.8\linewidth}
        \centering
        \includegraphics[width=\linewidth]{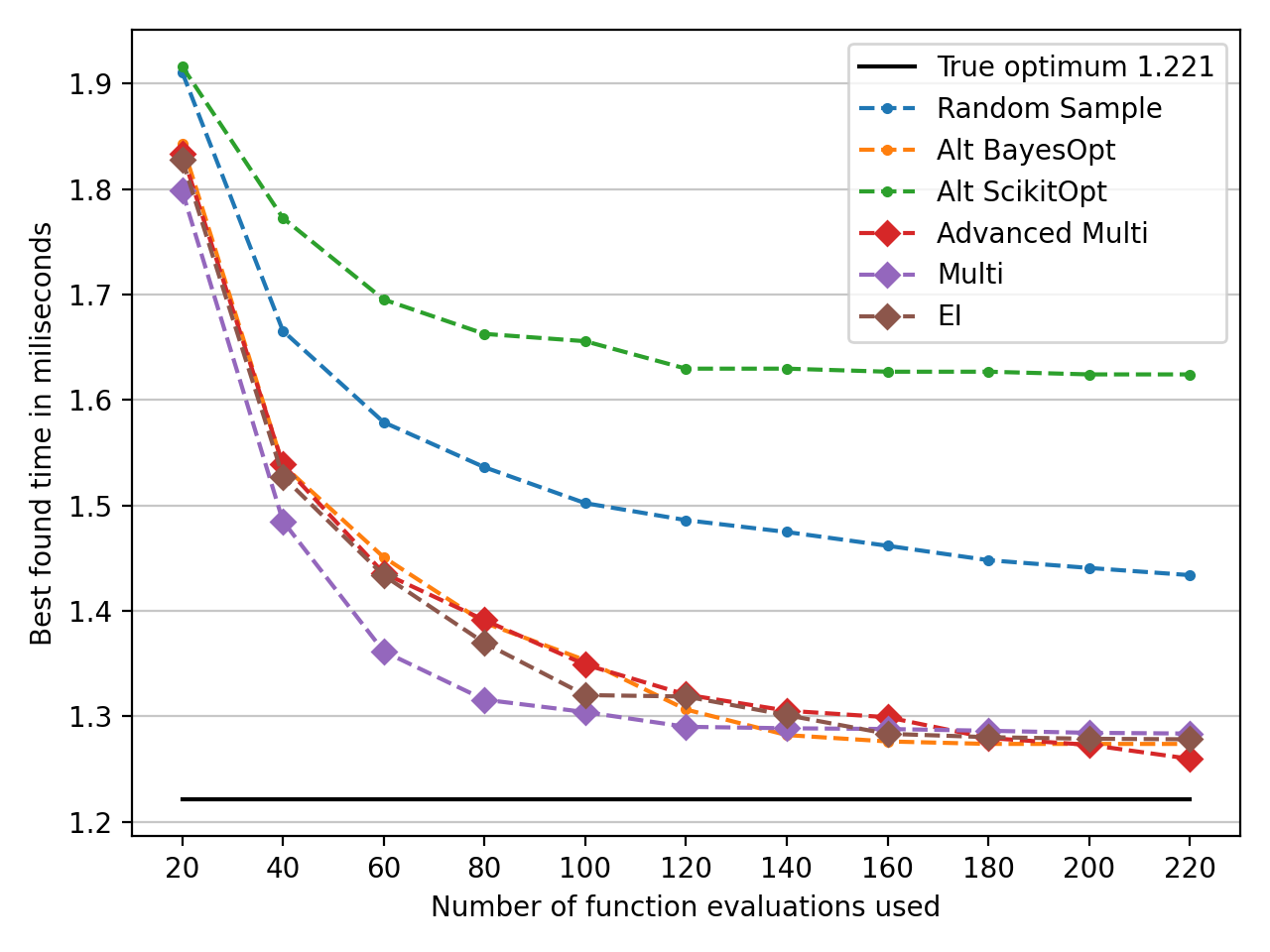}
        \caption{Convolution}
        \label{fig:results_frameworks_comparison_convolution}
    \end{subfigure}
    \begin{subfigure}[b]{0.8\linewidth}
        \centering
        \includegraphics[width=\linewidth]{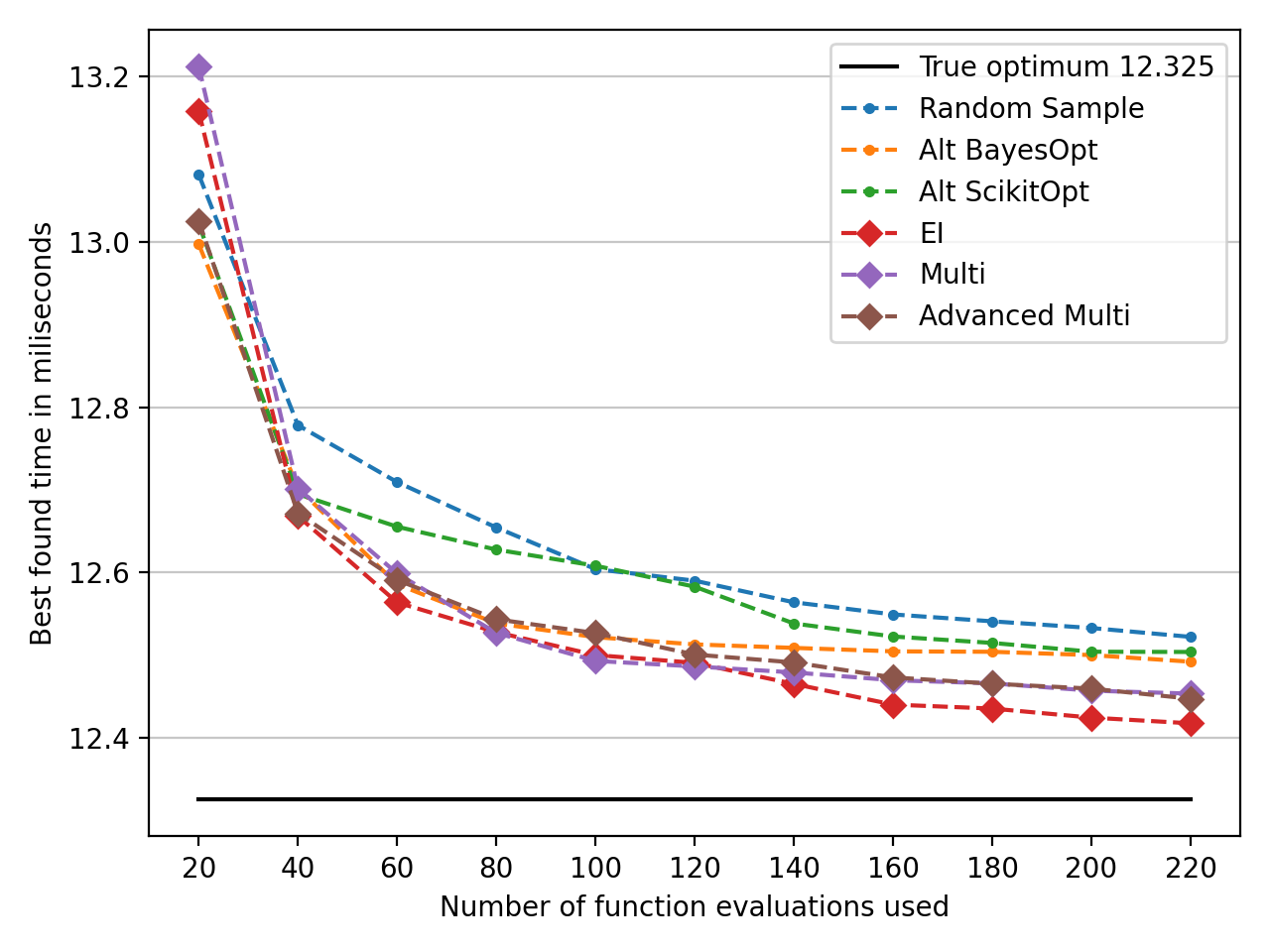}
        \caption{PnPoly}
        \label{fig:results_frameworks_comparison_pnpoly}
    \end{subfigure}
    \begin{subfigure}[b]{0.8\linewidth}
        \centering
        \includegraphics[width=\linewidth]{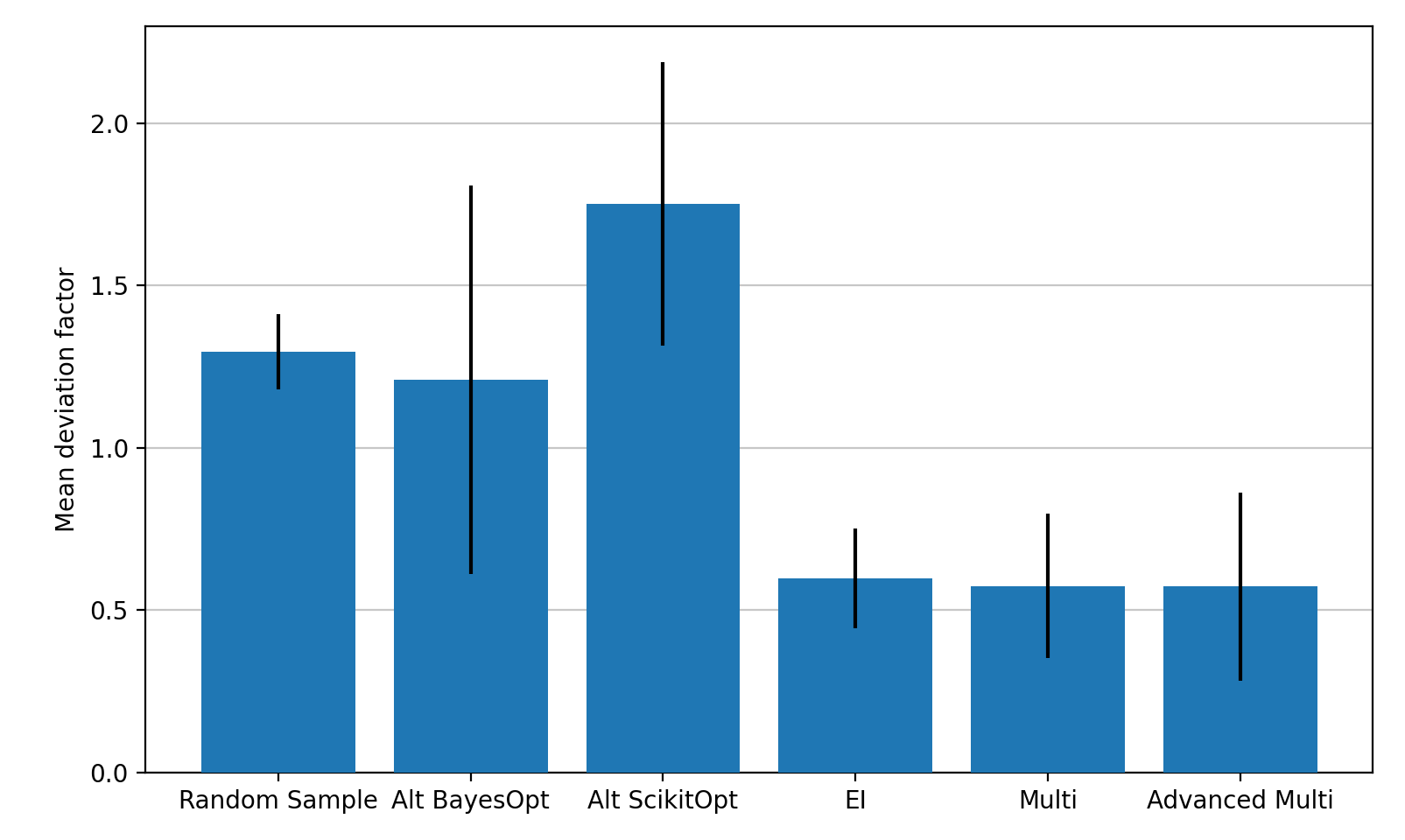}
        \caption{Mean deviation factors}
        \label{fig:results_frameworks_comparison_factors_gpu}
    \end{subfigure}
    \caption{Comparison with alternative Bayesian Optimization frameworks for various kernels with the RTX 2070 Super}
    \label{fig:results_frameworks_comparison}
\end{figure}

As seen in \cref{fig:results_frameworks_comparison_GEMM}, the BayesOpt and ScikitOpt implementations perform similar to each other on GEMM, noticeably performing worse than random search. 
The same holds for ScikitOpt on Convolution. 
This is likely because for GEMM and Convolution the search space is much smaller with the restrictions applied, which was not possible for these frameworks. 
However, even in PnPoly, which has no restrictions, performance of ScikitOpt is not much better than random search. 
\textit{BayesianOptimization} performs better than ScikitOpt on Convolution and PnPoly (\cref{fig:results_frameworks_comparison_convolution,fig:results_frameworks_comparison_pnpoly}). 
In Convolution, \textit{BayesianOptimization} even performs comparably to our methods. 
However, as seen in \cref{fig:results_frameworks_comparison_factors_gpu}, the general performance of these frameworks does not come close to our BO implementations.

\subsection{Unseen kernels}\label{sec:results-comparison-new-kernel}

We have consistently used the three GPU kernels, GEMM, Convolution, and PyPoly, for hyperparameter optimization and for performance comparisons. 
To verify that our strategies generalize well to other kernels, we also apply it to two previously unseen kernels, namely the ExpDist and the Adding kernels, executed on the NVIDIA A100. 

\begin{figure}[tbp]
    \centering
    \includegraphics[width=0.8\linewidth]{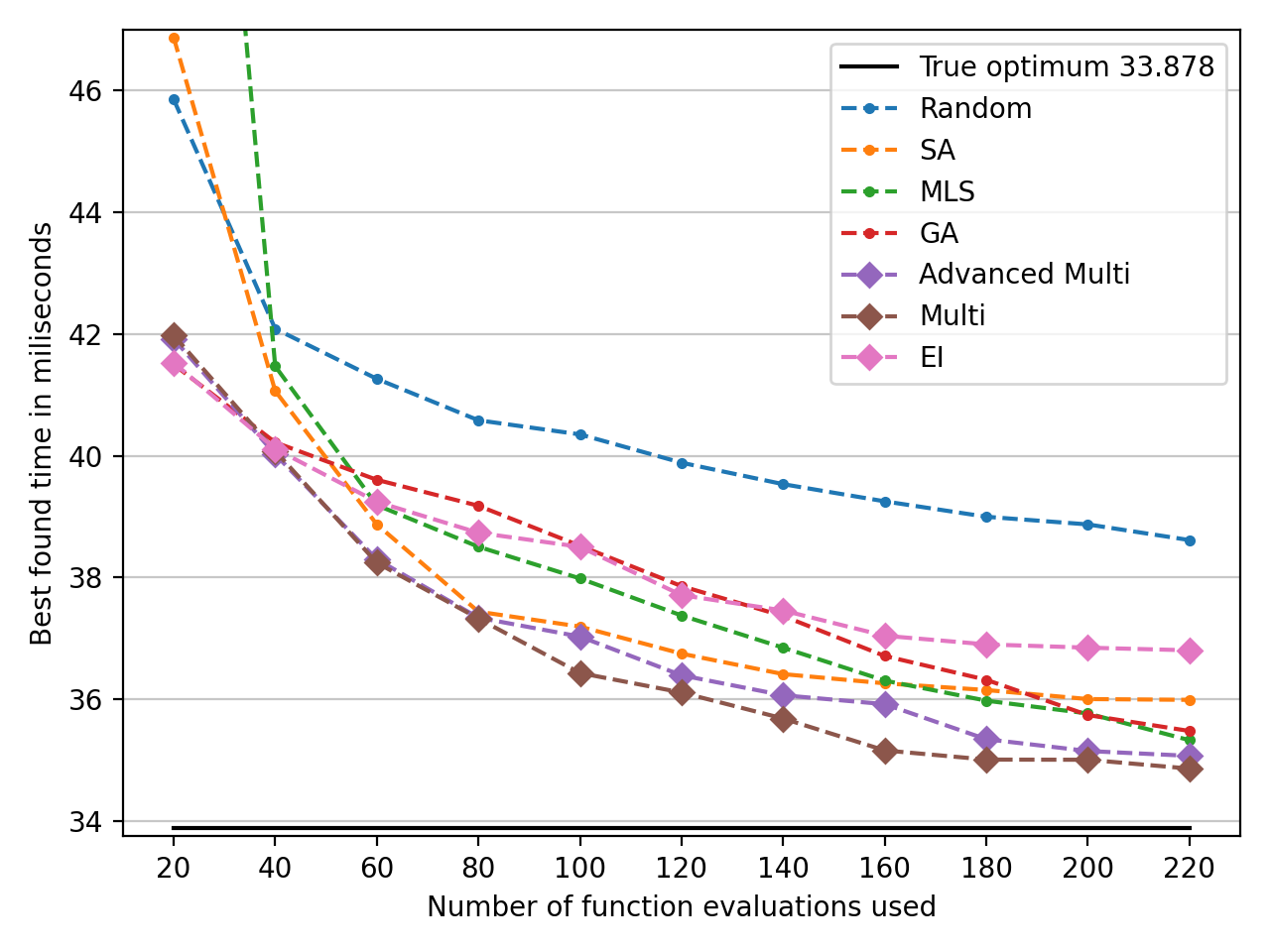}
    \vspace{-0.2cm}
    \caption{Performance on the ExpDist kernel with the A100}
    \label{fig:results_comparison_expdist}
\end{figure}

The ExpDist CUDA kernel is a double-precision GPU implementation of a simplified Bhattacharyya distance between two point sets that explicitly takes anisotropic localization uncertainty into account~\cite{expdist}. 
The tunable parameters describe the thread block dimensions, work per thread in two dimensions, partial loop unrolling factors, and the number of thread blocks to use in the y-dimension.
After applying constraints, the search space size consists of 14400 parameter configurations, of which 7320 (50.8\%) are invalid. 
The minimum in time lies at 12.764, however, the work executed by this kernel depends on the parameter configuration, so optimizing on time would result in the kernel that does the least work. Instead, we take $\frac{10^5}{{GFLOP/s}}$ as an alternative measure, with its minimum at 33.878. 
Observing \cref{fig:results_comparison_expdist}, \textit{advanced multi} and \textit{multi} perform very well. EI performs worse than the other Kernel Tuner methods and not nearly as well as the \textit{multi} strategies. 

\begin{figure}[tbp]
    \centering
    \includegraphics[width=0.8\linewidth]{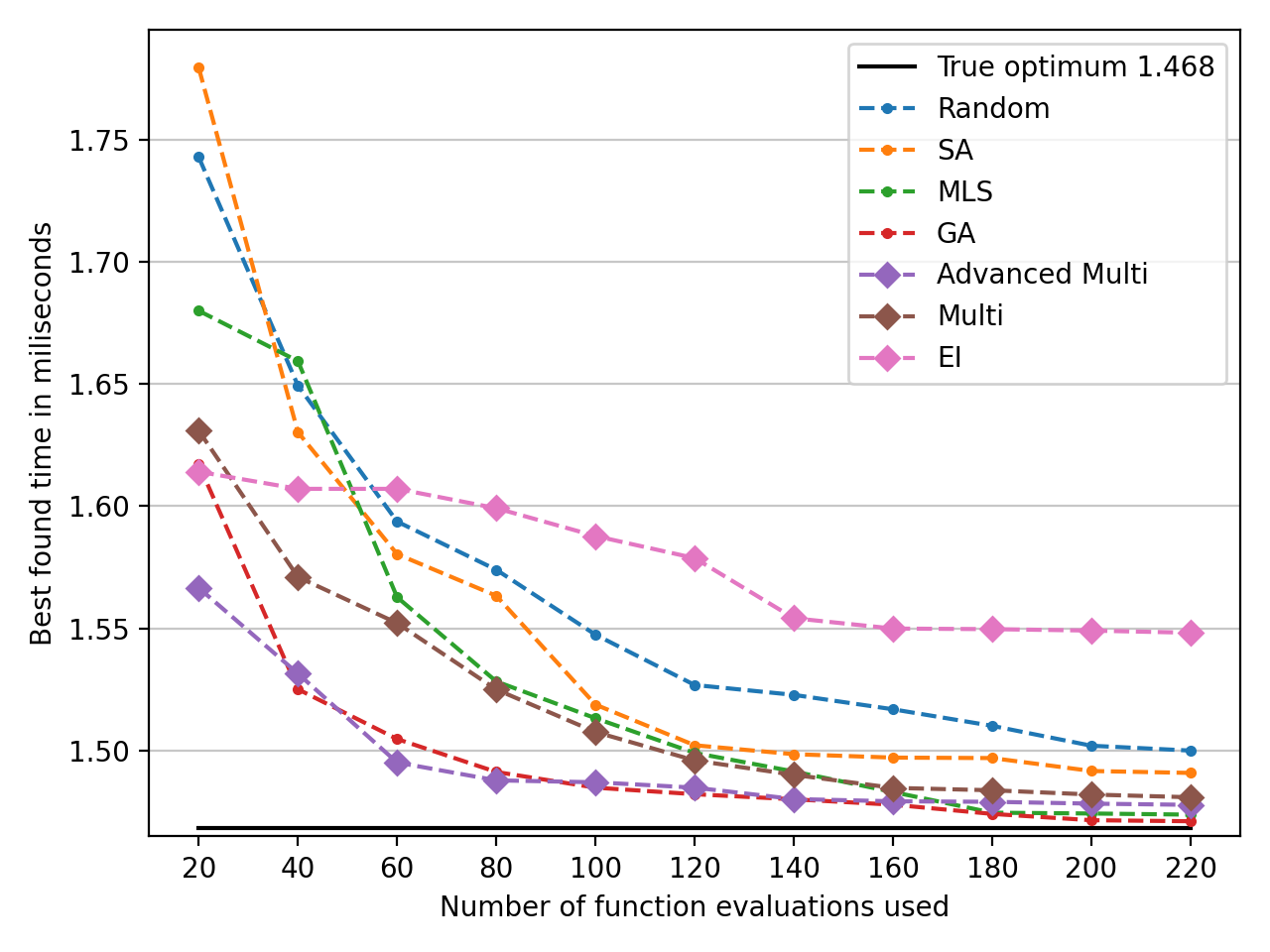}
    \vspace{-0.2cm}
    \caption{Performance on the Adding kernel with the A100}
    \label{fig:results_comparison_addingkernel}
\end{figure}

The Adding CUDA kernel computes the transport of diffuse radiation through a vertically layered atmosphere~\cite{adding_kernel}. The tunable parameters describe the thread block dimensions in x and y, a partial loop unrolling factor for the second loop in the kernel, which performs 140 iterations, and finally a switch whether or not to store a value computed in the first loop and load it again in the second loop, or to simply recompute that value in the second loop.
The search space after filtering is relatively small, consisting of just 4654 parameter configurations with none invalid, and a minimum of 1.468. 
In \cref{fig:results_comparison_addingkernel} it can be seen that EI appears unable to optimize well, performing worst of all search strategies. 
MLS, GA, \textit{advanced multi} and \textit{multi} appear to perform similarly to each other. 

\subsection{Discussion}

Across all kernels and GPUs, \textit{advanced multi} performed best in general, while the strongest competitor against our BO-methods in Kernel Tuner was a Genetic Algorithm. 
Looking at the mean deviation factors on the A100 (including the expdist and adding kernels), we find that \textit{advanced multi} performs 19.8\% better than GA. 
Moreover, on the GTX Titan X, \textit{advanced multi} performs 65.6\% better than GA, and on the RTX 2070 Super, \textit{advanced multi} performs 63.6\% better than GA.
Thus on average, \textit{advanced multi} performs 49.7\% better at finding parameter configurations than the best other method (GA) in Kernel Tuner. In addition, it performs 75\% better than the next-best other method, Simulated Annealing.

\section{Conclusions}\label{sec:conclusions}

An important aspect of this research has been the constraints of the problem domain, GPU auto-tuning. The challenging aspects are a combination of discrete search spaces, constraints, rough covariance functions, and invalid configurations, providing a setting in which Bayesian Optimization has not been applied before. 
To take on these problems, we implemented BO using a mapping to a discrete search space, with a Matérn covariance function, where parameter values are normalized and acquisition functions solely optimized on the non-evaluated parameter configurations. 
In addition, we propose a Contextual Variance function as an exploration factor to dynamically guide exploration, and propose the \textit{multi} and \textit{advanced multi} acquisition functions to improve scalability and make an informed selection of basic acquisition functions. Finally, we have optimized the hyperparameters of our method to arrive at a set of default parameters that should perform well for most tunable kernels.


To compare the performance of our methods, we used the GEMM, Convolution and PnPoly kernels on three different GPUs, and the ExpDist and Adding kernels on the A100 GPU. Our search strategies \emph{consistently} performed well, outperforming most existing search strategies in Kernel Tuner. Where some existing search strategies perform well on some test cases, they often perform relatively poor on others. 
On average over the mean deviation factors, \textit{advanced multi} performs 49.7\% better at finding parameter configurations than the best other method (GA) in Kernel Tuner, and 75\% better than the next-best other Kernel Tuner method (SA). 
We conclude that in general, \textit{advanced multi} is the best performing method, with a wide margin from other Kernel Tuner methods. 
Our methods outperform both the other search methods in Kernel Tuner and the alternative BO-frameworks. We conclude that it is not simply Bayesian Optimization in itself, nor the setup provided by Kernel Tuner, but our specific novel implementation that results in such a sizeable improvement over other methods. 

While we believe the presented methods are feasible for practical use, we have several ideas for further improvements. 
For example, by applying an optimizer on the acquisition function optimization, we plan to further reduce the run time.

\section*{Acknowledgment}
The CORTEX project has received funding from the Dutch Research Council (NWO) in the framework of the NWA-ORC Call (file number NWA.1160.18.316). 

\bibliographystyle{IEEEtran}
\bibliography{references,references-thesis}

\end{document}